\documentclass{article}

\PassOptionsToPackage{round}{natbib}

\usepackage[preprint]{neurips_2026}

\usepackage[utf8]{inputenc} 
\usepackage[T1]{fontenc}    
\usepackage[hidelinks]{hyperref}       
\usepackage{url}            
\usepackage{booktabs}       
\usepackage{amsfonts}       
\usepackage{nicefrac}       
\usepackage{microtype}      
\usepackage{xcolor}         
\usepackage{amsmath}
\usepackage{amsthm}
\usepackage{mathtools}
\usepackage{cleveref}
\usepackage{bm}

\newtheorem{proposition}{Proposition}

\usepackage{subcaption}
\usepackage{booktabs}
\usepackage[table]{xcolor}
\usepackage{wrapfig}

\title{Towards More General Control of Diffusion Models Using Jeffrey Guidance}

\author{%
  Raphaël~Razafindralambo\\
   Inria, CNRS, I3S, Maasai\\
   Université Côte d’Azur, France
   \\
  \texttt{raphael.razafindralambo@inria.fr} \\
  \And
  Rémy~Sun\\
  Inria, CNRS, I3S, Maasai\\
  Université Côte d’Azur, France\\
  \texttt{remy.sun@inria.fr} \\
  \And
  Frédéric~Precioso\\
  Inria, CNRS, I3S, Maasai\\
  Université Côte d’Azur, France\\
  \texttt{frederic.precioso@univ-cotedazur.fr} \\
  \And
  Jes~Frellsen\\
  Technical University of Denmark, Denmark\\
  \texttt{jefr@dtu.dk} \\
  \And
  Pierre-Alexandre~Mattei\\
  Inria, CNRS, LJAD, Maasai\\
  Université Côte d’Azur, France\\
  \texttt{pierre-alexandre.mattei@inria.fr} \\
}

\begin{document}

\maketitle

\begin{abstract}
A key strength of diffusion models lies in their flexibility, since their outputs can be controlled at sampling time through \emph{guidance}.  However, beyond simple cases such as conditional sampling, the target distribution is often left implicit, defined only through a sampling rule or a heuristic energy function. To address this, we propose Jeffrey guidance, a principled framework that extends diffusion-model control to applications beyond what standard guidance can express. It leverages Jeffrey’s rule of conditioning to update marginal distributions towards a prescribed target, preserving the conditional structure and minimally perturbing the joint distribution. We first demonstrate Jeffrey guidance by targeting a prescribed embedding distribution. With Inception embeddings as the target, this leads to substantial reductions in FID on both CIFAR-10 and FFHQ. We further apply Jeffrey guidance to fairness on CelebA-HQ, updating an unconditional diffusion model to enforce independence between attributes.
\end{abstract}

\section{Introduction}

Diffusion models \citep{sohl2015deep,ho2020denoisingdiffusionprobabilisticmodels,song2021scorebasedgenerativemodelingstochastic} have emerged as a powerful class of generative models producing high-quality images. They have achieved state-of-the-art performance across a wide range of applications, including image generation \citep{dhariwal2021diffusion}, text-to-image synthesis \citep{saharia2022photorealistic}, video modeling \citep{ho2022video}, molecular design \citep{liu2023generative}, and tabular data generation \citep{jolicoeur2024generating}.

A key strength of diffusion models lies in their flexibility, since their outputs can be controlled at sampling time through \emph{guidance} without retraining the underlying neural network. Just by adding a correction term to tilt the generation toward desired properties or constraints, guidance techniques enable conditioning on specific classes \citep{dhariwal2021diffusion,ho2022classifier}, with application to semantic editing \citep{brack2023sega} or debiasing for safety \citep{schramowski2023safe}. They also support matching a target distribution for fairness objectives \citep{parihar2024balancing}, or enforcing more complex criteria such as reducing training set memorization \citep{chen2024towards}. 

Beyond standard guidance methods such as class-conditional sampling \citep{dhariwal2021diffusion,ho2022classifier}, many approaches lack an explicitly defined target distribution. They instead often rely on heuristic modifications of the sampling dynamics and hyperparameter choices, notably in fairness \citep{kang2025fairgen,2025debiasingdiff}, concept control \citep{gandikota2024concept}, and compositional generation \citep{liu2022compositional}, effectively tilting the original distribution but making the actual target difficult to characterize explicitly or interpret.

\emph{Jeffrey's rule of conditioning} precisely provides a general recipe for updating probability distributions to satisfy explicit constraints. Introduced by \citet{jeffrey1957contributions} in the context of modeling belief updates in philosophy of science, it has been mostly studied in epistemology \citep{meehan2020jeffrey} with relatively limited adoption in statistics and machine learning. Given a global joint model, it allows updates of a marginal to a reference distribution while minimally modifying the original model. Notably, \citet{jeffrey1957contributions} points out standard conditioning, at the heart of class-conditional frameworks in diffusion \citep{dhariwal2021diffusion,ho2022classifier}, arises as a special case. 

\begin{figure}
    \centering
    \includegraphics[width=0.99\linewidth, trim={1.5cm 1.6cm 0.7cm 0.4cm}, clip]{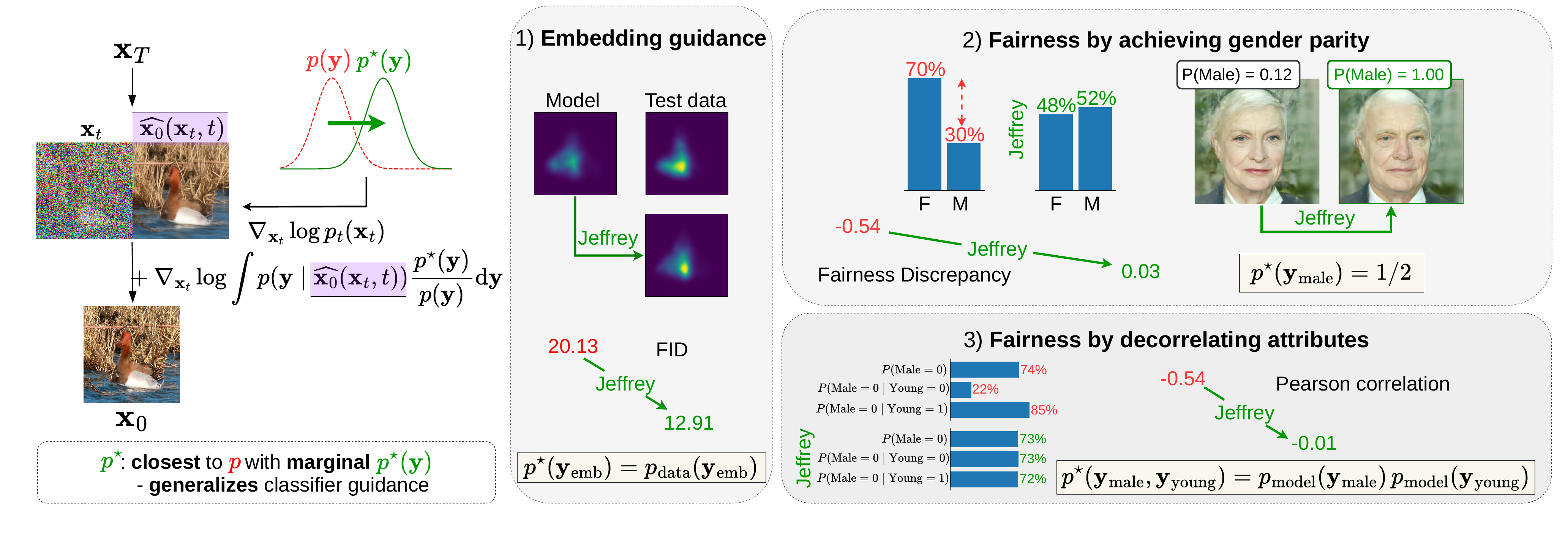}
    \caption{ \textbf{Overview of Jeffrey guidance.} While classifier guidance targets a single class, our method generalizes this and provides a distributional guidance framework that updates $p(\mathbf{y})$ toward a target $p^\star(\mathbf{y})$, with a correction term added at each diffusion step. This marginal-based framework successfully applies to various tasks, including embedding distribution matching and fairness objectives.}
    \label{fig:overview}
\end{figure}

In this paper, we take advantage of Jeffrey's rule to formulate a novel \emph{Jeffrey guidance} (see \Cref{fig:overview}). Unlike Bayes-based approaches \citep{dhariwal2021diffusion,ho2022classifier} whose goal is standard conditioning, we directly update the underlying distribution so that its marginals match a prescribed target. This allows us to address more general and complex objectives than standard conditioning, while maintaining precise and well-defined targets. Our approach still retains the simplicity of existing guidance methods, yielding a plug-and-play implementation that only requires adding a single term during sampling without retraining with the following main contributions:
\begin{itemize}
\item Jeffrey guidance, a more general diffusion guidance that uses Jeffrey's rule to match the generated output distribution to a target distribution. Jeffrey guidance both recontextualizes standard classifier guidance and opens up new possibilities for diffusion model control.
\item We match the distribution of model outputs to a target distribution in a given embedding space with Jeffrey guidance. As a proof of concept, we apply this to Inception embeddings to significantly lower model FID (without clear changes in generated images).
\item We control the distribution of image attributes with Jeffrey guidance. As a proof of concept, we show that CelebA attributes (\textit{Male} \& \textit{Young}) can be decorrelated in a Jeffrey guided model's outputs, leading to a fairer model.
\end{itemize}

\section{Preliminaries: Jeffrey’s rule and diffusion models}


\subsection{Jeffrey's rule of conditioning}


Bayes's rule, by allowing to condition on events or random variables, is the main tool that allows to perform probabilistic reasoning in machine learning. However, we would like in some cases to condition on more complex predicates than just events or random variables (e.g. to follow a certain embedding distribution). Jeffrey's rule \citeyearpar{jeffrey1957contributions} is a generalisation of Bayes's rule that allows to condition on such predicates. It is still a topic of active research in formal epistemology \citep{meehan2020jeffrey}, and a statistically-oriented review was written by \citet{diaconis1982updating}. It is yet barely discussed in machine learning, only sometimes in probablistic programming \citep{tolpin2021probabilistic,munk2023uncertain} or for protein structure prediction \citep{hamelryckalphafold}.

We present below a simple definition of Jeffrey's conditioning, based on density factorisations. More general measure-theoretic presentations are available in \citet[Section 6]{diaconis1982updating} and \citet{meehan2020jeffrey}.
We consider a joint density over a product space $\mathcal{X} \times \mathcal{Y}$,
\begin{equation}
p(x,y) = p(x \mid y)\, p(y),
\end{equation}
where $\mathcal{X}$ might denote the space of human face pictures, and $\mathcal{Y}$ corresponds to a set of associated features (e.g., gender, or age). More generally, $\mathcal{Y}$ may represent a collection of features $y = (y_1, \dots, y_d)$ on the product space $\mathcal{Y}_1 \times \cdots \times \mathcal{Y}_d$ so we can control multiple features simultaneously.

We aim to alter $p$ to modify the marginal distribution over $\mathcal{Y}$, replacing $p(y)$ with a target density $p^{\star}(y)$ (e.g. a desired distribution over attributes like gender parity), while preserving the conditional $p(x \mid y)$. Under this constraint, the \emph{Jeffrey-updated joint distribution} is defined as
\begin{equation}
p^{\star}(x,y) = p(x \mid y)\, p^{\star}(y).
\end{equation}
By design, the conditional distribution \mbox{$p^{\star}(x \mid y) = p(x \mid y)$} is preserved. This is the ``J-condition'' \citep{diaconis1982updating} or ``rigidity'' in epistemology \citep{meehan2020jeffrey}. However, this does not apply to other conditionals and marginals: for instance, the new marginal over $\mathcal{X}$ is \begin{equation}\label{eq: updated marginal}
p^{\star}(x) = \int p(x \mid y)\, p^{\star}(y)\,\mathrm{d}y
,\end{equation}
which is not necessarily equal to $p(x)$. Fortunately, $p$ and $p^\star$ are not too far away. Indeed, among all distributions $q$ that meet the constraint $q(y) = p^{\star}(y) $, $p^\star$ is exactly the closest one to $p$ \citep[Theorem 6.1]{diaconis1982updating}. Thus, the Jeffrey update may be interpreted as an information projection \citep{csiszar2003information} onto this set of distributions with prescribed marginal.

In our context, we will be particularly interested in expressing $p^{\star}(x)$ as a reweighted version of $p(x)$. If we assume that $p^\star(y)$ is absolutely continuous with respect to $p(y)$, then, starting from \Cref{eq: updated marginal} and using Bayes's rule we obtain
\begin{equation}\label{eq: alternative jeffrey}
p^{\star}(x) = p(x) \int p(y \mid x) \rho(y) \mathrm{d}y,
\quad \text{where } \rho(y) \coloneqq \frac{p^{\star}(y)}{p(y)}.
\end{equation}
An interesting case arises when the features are deterministically derived from $x$, i.e., $y = u(x)$ for some mapping $u: \mathcal{X} \rightarrow \mathcal{Y}$. In this setting, $p(y \mid x)$ becomes a point mass at $u(x)$ and \Cref{eq: alternative jeffrey} simplifies to the following reweighting,
\begin{equation}\label{eq: deterministic jeffrey}
p^{\star}(x) = \rho(u(x))\, p(x).
\end{equation}
Another particular case of interest is when $y$ is discrete and $p^\star(y)$ is a point mass at a certain point $c$. Then, $p^\star(x) = p(x | c)$, which means that standard conditioning is a particular case of Jeffrey conditioning. We will use this to frame classifier guidance as a particular case of Jeffrey guidance.

\subsection{SDE-based diffusion models and how to guide them}



Our goal will be to use Jeffrey's rule to control score-based diffusion models. We will rely on the fact that, starting from an initial diffusion model $p(\mathbf{x})$, it is possible to approximately sample from a reweighted model $p^\star(\mathbf{x})$ using guidance techniques.
Starting from a data distribution $p_0(\mathbf{x}_0) \coloneqq p(\mathbf{x}_0)$, an given a set of scalars $\{\alpha_t\}_{t=1}^T$, the forward process progressively perturbs the data into noise and yields:
\begin{equation}\label{eq: forward process}
\mathbf{x}_t = \sqrt{\alpha_t} \mathbf{x}_{t-1} + (\sqrt{1 - \alpha_t}) \mathbf{z} \quad \quad \mathbf{z} \sim \mathcal{N}(\boldsymbol{0},\boldsymbol{I}),
\end{equation}
where $\mathbf{z}$ and $t > 0$ \citep{song2021scorebasedgenerativemodelingstochastic,ho2020denoisingdiffusionprobabilisticmodels}. We define for each time $t>0$ the density $p_t(\mathbf{x}_t) =
\int p_{t\mid 0}(\mathbf{x}_t \mid \mathbf{x}_0) p_0(\mathbf{x}_0)\, \mathrm{d}\mathbf{x}_0$, obtained by noising the initial distribution $p_0(\mathbf{x}_0)$ through the transition kernel $p_{t\mid 0}(\mathbf{x}_t \mid \mathbf{x}_0)$ given by \Cref{eq: forward process}. Starting from Gaussian noise $\mathbf{x}_T$, one can then sample from $p(\mathbf{x})$ by simulating the corresponding backward process that depends on a time-dependent score \citep{song2021scorebasedgenerativemodelingstochastic},
\begin{equation}\label{eq: backward process}
\mathbf{x}_{t-1}=\frac{1}{\sqrt{\alpha_t}}\left(\mathbf{x}_t+(1-\alpha_t)\underbrace{\nabla_{\mathbf{x}_t}\log p_t(\mathbf{x}_t)}_{\text{score}}\right)+\sigma_t\mathbf{z},\qquad \mathbf{z}\sim\mathcal{N}(0,\mathbf{I}),
\end{equation} 
where $\sigma_t > 0$. Using the denoising score matching objective \citep{song2021scorebasedgenerativemodelingstochastic}, a neural network $\boldsymbol{s}_{\bm{\theta}}(\mathbf{x}_t,t)$ is trained to estimate $\nabla_{\mathbf{x}_t} \log p_t(\mathbf{x}_t)$ for each $t>0$. 


Given a pre-trained model $p(\mathbf{x})$, it is possible to sample from a modified target distribution $p^{\star}(\mathbf{x})$ obtained by reweighting $p(\mathbf{x})$ with a multiplicative factor. A canonical example is class-conditional generation, where the goal is to sample from
$p^{\star}(\mathbf{x}) \coloneqq p(\mathbf{x} \mid c) \propto p(\mathbf{x})\, p(c \mid \mathbf{x})$
with $c$ being the target class \citep{dhariwal2021diffusion}. More generally, guidance techniques provide ways of sampling from a reweighted distribution
\begin{equation}\label{eq: energy formula}
p^{\star}(\mathbf{x}) \propto p(\mathbf{x}) \exp(-\lambda \mathcal{E}(\mathbf{x})),
\end{equation}
where $\lambda > 0$ and $\mathcal{E}: \mathcal{X} \to \mathbb{R}$ is an energy function. To sample from $p^{\star}$ using a diffusion process, we need some estimate of the score
$\nabla_{\mathbf{x}_t} \log p_t^{\star}(\mathbf{x}_t)$ 
with $p_t^{\star}(\mathbf{x}_t) = \int p_{t\mid 0}(\mathbf{x}_t \mid \mathbf{x}_0)\, p^{\star}(\mathbf{x}_0)\, \mathrm{d}\mathbf{x}_0$ for each $t>0$. In practice, one exploits \Cref{eq: energy formula} to define
\begin{equation}\label{eq: guidance energy formula}
\nabla_{\mathbf{x}_t} \log \tilde{p}^{\star}_t(\mathbf{x}_t)
\coloneqq \nabla_{\mathbf{x}_t} \log p_t(\mathbf{x}_t)
- \lambda \nabla_{\mathbf{x}_t} \mathcal{E}_t(\mathbf{x}_t),
\end{equation}
where $\nabla_{\mathbf{x}_t} \log p_t(\mathbf{x}_t)$ is estimated from the pre-trained model $\boldsymbol{s}_{\bm{\theta}}(\mathbf{x}_t,t)$, and $\mathcal{E}_t$ is a time-dependent energy approximated using standard techniques. An approach (\citealp{chung2022diffusion,bansal2023universal}, \textit{universal guidance}) consists in using a time-dependent inference $\widehat{\mathbf{x}}_0(\mathbf{x}_t,t) = \mathbb{E}_{p_{t \mid 0}(\mathbf{x}_0 \mid \mathbf{x}_t)}[\mathbf{x}_0]$ of the clean sample $\mathbf{x}_0$ obtained from $\mathbf{x}_t$, and plugging it into the energy:\begin{equation}
\widehat{\mathcal{E}}_t(\mathbf{x}_t) \coloneqq \mathcal{E}(\widehat{\mathbf{x}}_0(\mathbf{x}_t,t)).
\end{equation}
In our previous example of a class-conditional setting, this would correspond to computing $\widehat{\mathcal{E}}_t(\mathbf{x}_t) = -\log p(c \mid \widehat{\mathbf{x}}_0(\mathbf{x}_t,t))$ at each step. Although some approaches learn time-dependent energy functions defined directly on noisy samples $\mathbf{x}_t$ and $t$ without $\widehat{\mathbf{x}}_0$ \citep{ho2022classifier,lu2023contrastive}, they incur a higher computational cost due to the need for larger neural networks. Using $\widehat{\mathbf{x}}_0$ avoids this dependence on time and reduces the overhead \citep{bansal2023universal} since one learns the energy on clean samples, although it results in an approximate sampling procedure. Due to inaccurate $\widehat{\mathbf{x}}_0$ predictions at early timesteps, guidance can be restricted to a window of discrete timesteps, e.g. $\{1,2,\dots,\delta\}$ with $\delta<T$, during reverse diffusion sampling \citep{2025debiasingdiff,chen2024towards}.

\section{Jeffrey guidance and its applications}
\label{sec:jeffrey_guidance}

We present Jeffrey guidance (see \Cref{sec:jg_density}), a novel framework that leverages Jeffrey's rule to guide diffusion to match a prescribed target, as well as two new applications it opens up: embedding guidance (see \Cref{sec:embedding}) and finer attribute distribution control (see \Cref{sec:fairness}).

\subsection{Jeffrey guidance: a density ratio-based framework}
\label{sec:jg_density}

In this section, we show how Jeffrey's rule of conditioning, with the goal of updating a marginal distribution to a prescribed, can be naturally incorporated into the guidance framework.
We define the product space $\mathcal{X} \times \mathcal{Y}$, where $\mathcal{X} \subset \mathbb{R}^{d}$ is the image space and $\mathcal{Y} = \mathcal{Y}_1 \times \cdots  \times \mathcal{Y}_n$ is a space of attributes/features. We start with an initial joint model $p(\mathbf{x},\mathbf{y})$, where $p(\mathbf{x})$ has been approximated by a pre-trained diffusion model. Our goal is to alter the model in order to match a marginal $p^\star(\mathbf{y})$. Using \eqref{eq: updated marginal}, the Jeffey update will satisfy
\begin{equation}\label{eq: tilted target distribution jeffrey diffusion}
p^{\star}_0(\mathbf{x})  =  p_0(\mathbf{x}) \left(\int p(\mathbf{y} \mid \mathbf{x})\rho(\mathbf{y})\mathrm{d}\mathbf{y} \right),
\end{equation}
where $\rho(\mathbf{y}) = p^{\star}(\mathbf{y})/p(\mathbf{y})$. We recognize \Cref{eq: energy formula} with $\mathcal{E}(\mathbf{x}) = -\log\int p(\mathbf{y} \mid \mathbf{x})\rho(\mathbf{y})\mathrm{d}\mathbf{y}$. 

Following the derivation in \Cref{eq: guidance energy formula}, we can therefore define the guided score as
\begin{equation}\label{eq: guided score x0 hat}
\nabla_{\mathbf{x}_t} \log \tilde{p}^{\star}_t(\mathbf{x}_t)
= \nabla_{\mathbf{x}_t} \log p_t(\mathbf{x}_t)
+ \nabla_{\mathbf{x}_t} \log  \int p(\mathbf{y} \mid \widehat{\mathbf{x}}_0(\mathbf{x}_t,t))\rho(\mathbf{y})\mathrm{d}\mathbf{y}.
\end{equation}
 The first term is obtained from the pre-trained score model. The second term requires estimating both the density ratio and the feature mapping. Generally, and especially for tasks where $\mathbf{y}$ is discrete, $\rho(\mathbf{y})$ can be derived upstream of the reverse diffusion process for each $\mathbf{y}$. When $\rho(\mathbf{y}) = \delta_{c}(\mathbf{y})/p_0(y)$ where $c$ is a given class, this additional term reduces to a variant of class-conditional guidance \citep{ho2022classifier} that would use $\widehat{\mathbf{x}}_0$.

We get $\widehat{\mathbf{x}}_0$ from the reverse inference rule (\citealp{efron2011tweedie}, \textit{Tweedie's formula}) with $\bar{\alpha}_t \coloneqq \prod_{i=1}^T \alpha_i$ and
\begin{equation}
\widehat{\mathbf{x}_0}(\mathbf{x}_t,t) =\frac{1}{\sqrt{\bar{\alpha}_t}} 
\left(\mathbf{x}_t+(1-\bar{\alpha}_t)\nabla_{\mathbf{x}_t}\log p_t(\mathbf{x}_t)\right)\approx \frac{1}{\sqrt{\bar{\alpha}_t}} 
\left(\mathbf{x}_t+(1-\bar{\alpha}_t)\boldsymbol{s}_{\bm{\theta}}(\mathbf{x}_t,t)\right).
\end{equation}

In the case where $\mathbf{y}$ is deterministically given by $\mathbf{x}$, namely for each data point $\mathbf{y} = u(\mathbf{x})$ where $u: \mathcal{X} \rightarrow \mathcal{Y}$ is deterministic, we can use \Cref{eq: deterministic jeffrey} and consider the tilted target distribution
\begin{equation}
p^{\star}_0(\mathbf{x}) =  p_0(\mathbf{x}) \rho(u(\mathbf{x})).
\end{equation} 
In this case, we identify $\mathcal{E}(\mathbf{x}) = - \log \rho(u(\mathbf{x}))$ and we define the guided score
\begin{equation}\label{eq: guided score x0 hat deterministic}
\nabla_{\mathbf{x}_t} \log \tilde{p}^{\star}_t(\mathbf{x}_t)
= \nabla_{\mathbf{x}_t} \log p_t(\mathbf{x}_t)
+ \nabla_{\mathbf{x}_t}\log \rho(u( \widehat{\mathbf{x}}_0(\mathbf{x}_t,t)).
\end{equation}
Despite its simplicity, this formula still requires $u$ to be differentiable. We note that in both cases the score at time $t>0$ remains approximated and we do not actually follow the path induced by $p^\star_t(\mathbf{x}_t)$. Exactness can however be recovered through more complex methods (see Appendix~\ref{app: exactness}).

In practice, we set up an additional hyperparameter $\lambda>0$ to scale the new correction term of the score. It is commonly used to control the trade-off between sample quality and faithfulness to the target distribution, notably in text- or class-conditional settings \citep{dhariwal2021diffusion,ho2022classifier,rombach2022high}. In a general perspective it is an inverse temperature controlling the energy strength (\citealp{lu2023contrastive}, \Cref{eq: energy formula}). When $\lambda = 1$ we strictly update according to Jeffrey's rule, but when $\lambda \neq 1$ it is not the exactly the case. 
\subsection{Enabling embedding guidance with Jeffrey guidance applied to Inception embeddings}
\label{sec:embedding}

We show here that Jeffrey guidance empowers us to update effectively an embedding distribution toward a reference one, something difficult to formulate with standard guidance. An interesting example is the Inception space, for which we describe here how to update the embedding distribution toward the training one. As a result, this procedure directly lowers of the Fréchet Inception Distance (FID), a metric widely used as a proxy evaluating generative image models  \citep{heusel2017gans}. 

In this case, $\mathbf{x}$ is an image and  $\mathbf{y} = u_{\text{in}}(\mathbf{x})$ is its corresponding Inception vector used to compute the FID. Let $q_{\text{train}}$ denote the training data distribution and $q_{\bm{\theta}}$ denote the generative model. The FID compares the marginals $q_{\bm{\theta}}(\mathbf{y})$ and $q_{\text{train}}(\mathbf{y})$, which are both unknown and only accessible through samples. However, this is sufficient in our Jeffrey settings since we will be able to estimate the density ratio from these samples.

We want to update the marginal on $\mathbf{y}$ to the target distribution $p^{\star}(\mathbf{y}) = q_{\text{data}}(\mathbf{y})$. 
According to \Cref{eq: tilted target distribution jeffrey diffusion}, this corresponds to reweighting $p(\mathbf{x})$ using the density ratio
\begin{equation}
\rho(\mathbf{y}) = \frac{p^{\star}(\mathbf{y})}{p(\mathbf{y})} = \frac{q_{\text{data}}(\mathbf{y})}{q_\theta(\mathbf{y})} = \frac{q_{\text{data}}(u_{\text{in}}(\mathbf{x}))}{q_\theta(u_{\text{in}}(\mathbf{x}))}.
\end{equation}

The crux of the issue therefore lies in estimating this ratio, which we chose to do directly via the standard technique of probabilistic classification \citep{sugiyama2012density}, that we detail in Appendix \ref{sec:density_ratio_estimation}. Specifically, we train a binary classifier to distinguish samples from $p^{\star}$ and $p$. Using Bayes' rule, the density ratio can then be expressed in terms of the optimal classifier.
Given this differentiable ratio estimator, since $u_{\text{in}}$ fortunately is differentiable, one can exploit \Cref{eq: guided score x0 hat deterministic} to apply guidance.

\subsection{Enabling attribute distribution matching and decorrelation applied to fairness}
\label{sec:fairness}

In this section, we show that Jeffrey guidance is able to handle distributional guidance in a general way. To illustrate this we tackle fairness problems, which amounts in our setting to controlling the distribution of socially sensitive attributes in generated human face images. In particular, we consider distribution matching, which enforces a desired marginal over attributes (e.g., gender balance), and decorrelation, which are tasks that standard classifier guidance cannot answer to straightforwardly. 

For context, existing works in fairness for diffusion models typically tackles statistical parity or targeting a fair proportion \citep{choi2020fair,kang2025fairgen,parihar2024balancing,friedrich2023fair}.
Beyond just standard guidance, \cite{parihar2024balancing} propose a post-hoc debiasing framework for diffusion models that minimizes a  \textit{Chi-square} distance between generated attribute distributions to a target distribution $p_{\mathrm{ref}}^a$ through an attribute predictor and batch-level guidance. \cite{2025debiasingdiff} propose a training-free and inference-time framework that is able to perform guidance using text and image modalities, in order to target a fair distribution. However, they do not extend their analysis to decorrelation of attributes, which is an important question in fairness. Moreover, while these methods optimize explicit matching objectives, the guidance procedure is not constructed to exactly target the desired distribution by design.

We consider the product space $\mathcal{X} \times \mathcal{A}$ where $\mathcal{A}$ is a product of (sensitive) attribute spaces (e.g. if $\mathcal{X}$ is the space of human face images, $\mathcal{A}$ can include genre, age, ...). We consider the following joint distribution $p(\mathbf{x},\mathbf{a}_1,\dots,\mathbf{a}_n) \coloneqq q_{\bm{\theta}}(\mathbf{x},\mathbf{a}_1,\dots,\mathbf{a}_n)$. We find it more appropriate for our debiasing task to address the bias of the model rather than that of the training set, since their attribute distributions may differ. In this case, one would instead consider $p=q_{\mathrm{data}}$.

\paragraph{Distribution matching.} To illustrate that Jeffrey guidance can steer toward a whole attribute distribution generalizing classifier guidance, we target a distribution $p^\star(\mathbf{a}_1)$ on a single attribute $\mathbf{a}_1$ with discrete values. The ratio of interest would be
\begin{equation}
\rho(\mathbf{a}_1)=\frac{p^\star(\mathbf{a}_1)}{p(\mathbf{a}_1)}
=\frac{p_{\mathbf{a}_1}}
{
q_{\bm{\theta}}(\mathbf{a}_1)
},
\end{equation}
where $p_{\mathbf{a}_1}$ denotes the target probability associated with the discrete value $\mathbf{a}_1$. In our experiments (see \Cref{sec:Parity_exp}), we consider a balanced target distribution, namely $p^\star(\mathbf{a}_1)=1/2$.

\paragraph{Decorrelation.} While enforcing attribute independence is difficult to express through standard classifier guidance, Jeffrey guidance naturally formulates it as a marginal to target. Indeed, \textit{mutual independence} is formulated as $p(\mathbf{a}_1,\mathbf{a}_2,\dots,\mathbf{a}_n) = p(\mathbf{a}_1)p(\mathbf{a}_2)\dots p(\mathbf{a}_n)$. As a result, $p^\star(\mathbf{a}_1,\dots,\mathbf{a}_n)$ can fortunately be identified easily and the ratio of interest is
\begin{equation}\label{eq: decorrelation ratio}
\rho(\mathbf{a}_1,\dots,\mathbf{a}_n) = \frac{p^{\star}(\mathbf{a}_1,\dots,\mathbf{a}_n)}{p(\mathbf{a}_1,\dots,\mathbf{a}_n)} = \frac{q_{\bm{\theta}}(\mathbf{a}_1)q_{\bm{\theta}}(\mathbf{a}_2)\dots q_{\bm{\theta}}(\mathbf{a}_n)}{q_{\bm{\theta}}(\mathbf{a}_1,\mathbf{a}_2,\dots,\mathbf{a}_n)}.
\end{equation}
In our experiments, we will consider the decorrelation of two discrete variables (see \Cref{sec:decorrelation_exp}).

In both cases, the ratio can be estimated using a classifier that learns to jointly predict the attributes (see Appendix \ref{subsec:appendix_decorrelation_attributes}).  Since the attributes are not deterministic nor differentiable with respect to the data, we rely on the base guidance term from \Cref{eq: guided score x0 hat}.

\section{Experiments}

\label{sec:experiments}


We evaluate our guidance framework on real-world image datasets spanning multiple resolutions. This includes low-resolution datasets such as CIFAR-10 \citep{krizhevsky2009learning}, consisting of $32 \times 32$ images, as well as higher-resolution datasets such as FFHQ \citep{karras2019style} and CelebA-HQ \citep{karras2017progressive}, both at $256 \times 256$ resolution, on which our diffusion models are pre-trained. For the fairness experiments conducted on CelebA-HQ, we additionally leverage the original CelebA dataset \citep{liu2015deep} to train the ratio estimator and the attribute predictors. 
All diffusion models use variants of the U-Net architecture \citep{ronneberger2015u}, that we detail in Appendix \ref{sec:experimental_details_appendix}. For ratio estimation, we use logistic regression implemented as a one-hidden-layer MLP classifier. This simple architecture is sufficient in our setting, as the inputs $\mathbf{y}$ to the classifier are essentially low-dimensional feature vectors (e.g., Inception embeddings) rather than high-dimensional images.
We adopt DDIM \citep{song2020denoising} with $100$ steps and $\eta = 0.2$ unless stated otherwise, and otherwise follow the default settings of the original work. When using the guided score in \Cref{eq: guided score x0 hat}, which relies on an approximation of $\mathbf{x}_0$, we introduce a hyperparameter $\delta$ that controls the starting point of the guidance. Specifically, since the estimation of $\widehat{\mathbf{x}}_0$ is highly noisy at early timesteps, guidance is only applied for timesteps $t \leq \delta$, where $\delta \leq 1000$.

We will compare Jeffrey guidance  against a simple baseline that also allows to target $p^\star$, and that combines ancestral sampling with standard guidance: first, we sample $y \sim p^\star$, then we use guidance to sample approximatively from $p(\mathbf{x} | \mathbf{y})$. We call this baseline ‘‘standard guidance''.

\subsection{Matching embedding distributions with Jeffrey guidance}
\label{sec:embedding_exp}

As outlined in \Cref{sec:embedding}, our Jeffrey guidance's more general formulation empowers us to match the distributions of generated images to a reference distribution in some embedding space. We illustrate this behavior by matching the Inception embedding distribution of generated images to that of the training set, which has clear effects reflected both in the FID metric (see \Cref{sec:FID_exp}) and the distribution of the embeddings themselves (see \Cref{sec:PCA_exp}). Interestingly, a detailed analysis in Appendix \Cref{app:FID_analysis} on the effect of Jeffrey guidance on generated images shows FID can lower drastically with little discernible changes in generated images which highlights limitations of FID as a perceptual metric. This is in line with the criticism of the FID led by \citet{jayasumana2024rethinking}. 

\subsubsection{Jeffrey guidance on Inception embeddings directly lowers FID}
\label{sec:FID_exp}

\begin{figure}[t]
    \centering
    \begin{subfigure}{0.32\linewidth}
        \includegraphics[width=\linewidth]{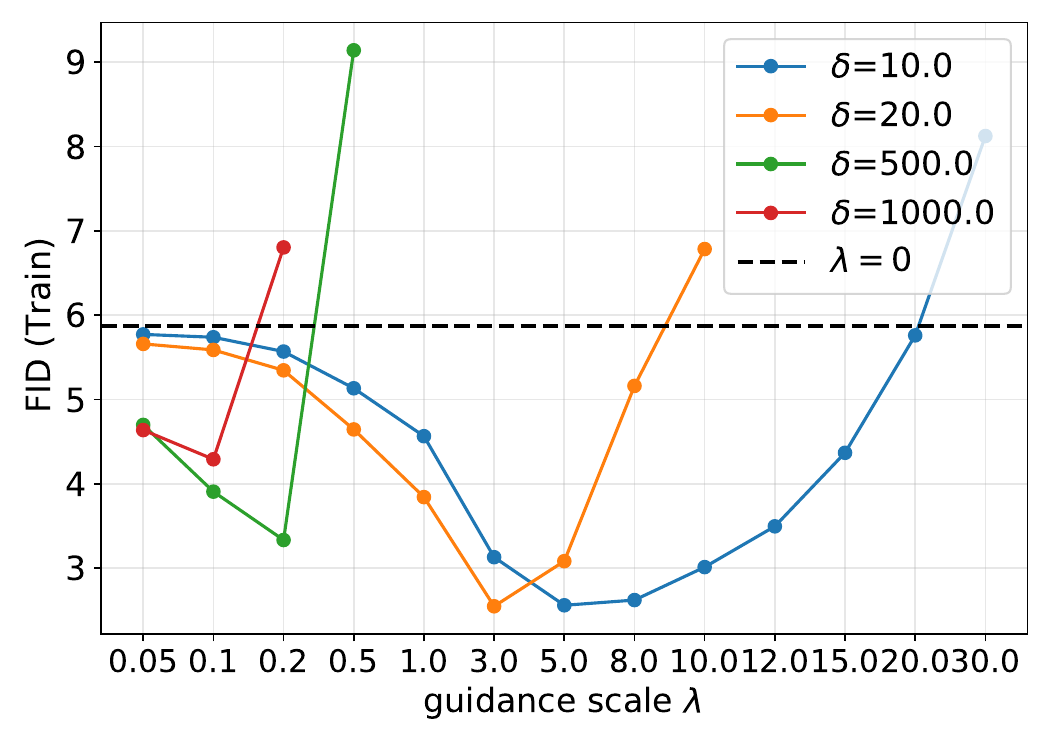}
        \caption{CIFAR-10 (train, 50k)}
    \end{subfigure}%
    \begin{subfigure}{0.32\linewidth}
        \includegraphics[width=\linewidth]{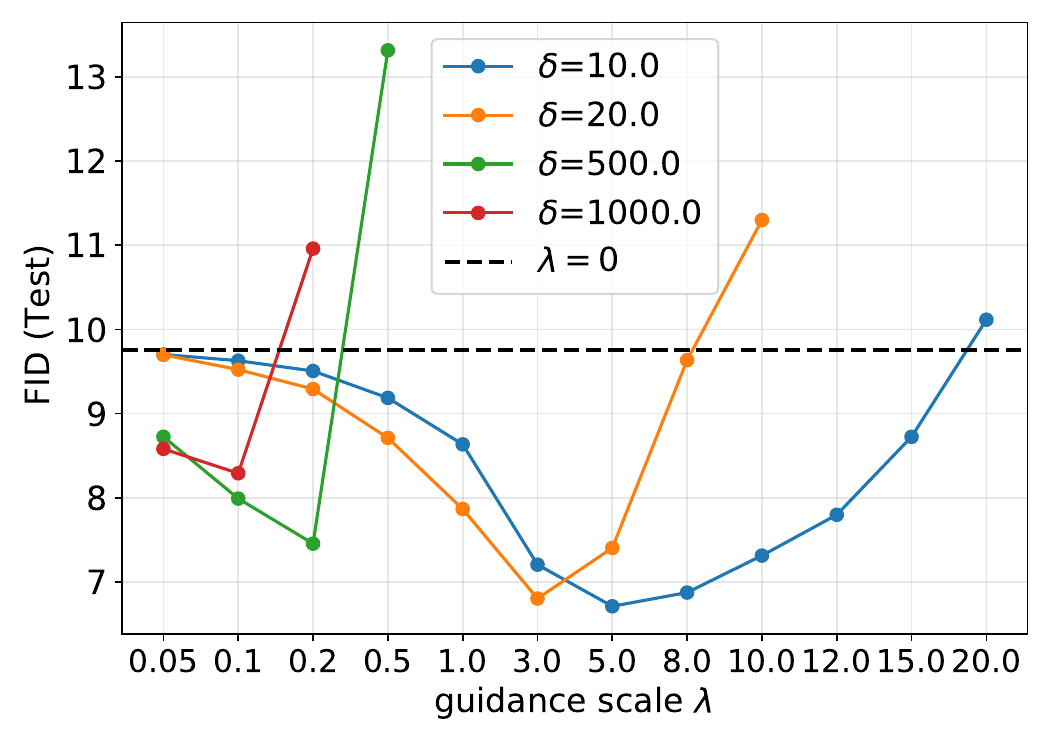}
        \caption{CIFAR-10 (test, 10k)}
    \end{subfigure}%
    \begin{subfigure}{0.32\linewidth}
        \includegraphics[width=\linewidth]{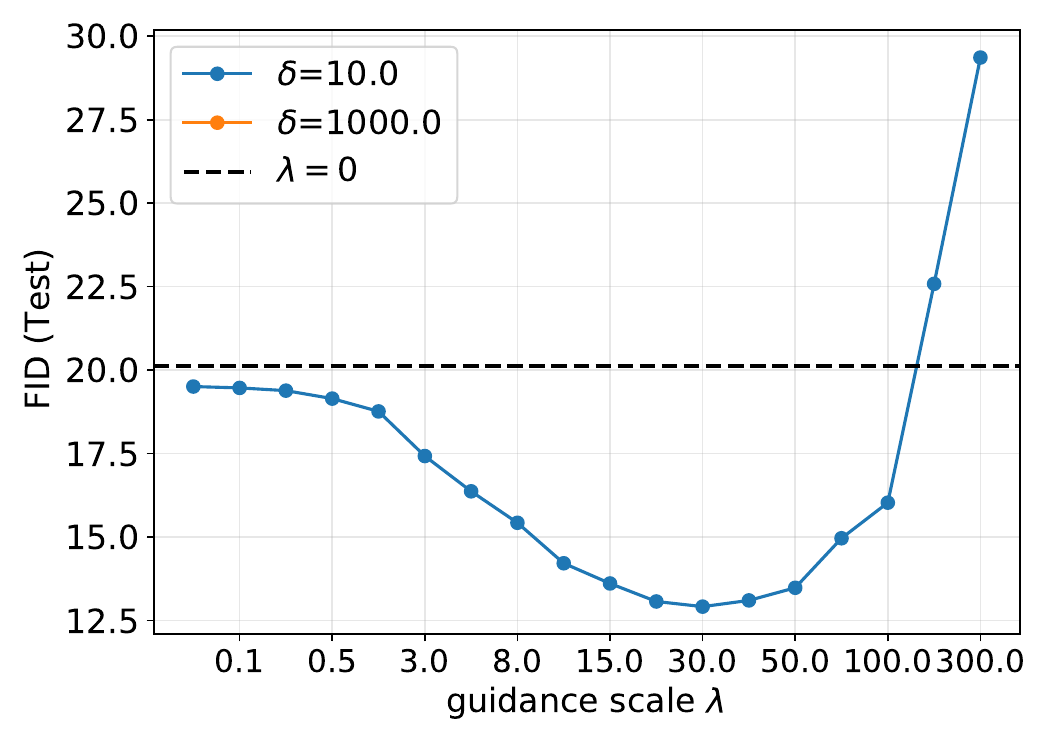}
        \caption{FFHQ (test, 10k)}
    \end{subfigure}
    \caption{
    \textbf{FID as a function of the guidance scale $\lambda$ for different values of $\delta$. }
    On CIFAR-10, guidance improves over the unguided baseline ($\lambda=0$) on both train and test sets for appropriate choices of $(\lambda, \delta)$. 
    We observe that larger values of $\delta$, corresponding to guidance applied earlier in the diffusion process through $\widehat{\mathbf{x}}_0$, require smaller guidance scales to remain stable, whereas smaller $\delta$ allows for stronger guidance as it is applied closer to the end of the trajectory. 
    Similar trends are observed on FFHQ, showing that the behavior extends to higher-resolution data.
    }
    \label{fig:jeffrey_fid}
\end{figure}

We first check that Jeffrey guidance successfully modifies the distribution of generated images' Inception embeddings by monitoring the resulting FID on CIFAR-10 and FFHQ. More precisely, we show that the generated samples exhibit a closer alignment with \textbf{both} the training and test distributions in the Inception embedding space. We stress that the density ratio estimator was trained on training embedding vectors and generated samples only (i.e. without ever seeing the test set). 

\Cref{fig:jeffrey_fid} shows FID notably improves over the baseline ($\lambda=0$) for all values of $\delta$ with moderate guidance scale $\lambda$ before eventually worsening as it becomes too strong and causes distortions. On CIFAR-10, the FID decreases from $\approx 5.88$ to $\approx 2.55$ on the training set (50k), and from $\approx 9.76$ to $\approx 6.71$ on the test set (10k). While our standard guidance baseline shows similar trends (see Appendix \Cref{fig:jeffrey_baseline_fid}), the gains are much lower compared to our method. On FFHQ (test set), the FID further drops from $\approx 20.13$ to $\approx 12.91$, indicating even larger gains. This significant drop confirms Jeffrey guidance can help match the distribution of high-dimensional embeddings of the data.

Interestingly, \Cref{fig:jeffrey_fid} suggests that lower $\delta$ values actually lead to better FID scores and allow for higher guidance scales $\lambda$ before worsening performance. As such, the cheapest option of only applying Jeffrey guidance to the last denoising step of the diffusion model ($\delta=10$) appears to be the most desirable configuration. We believe this interesting property is due to the fact Inception embeddings are computed on final denoised images and high $\delta$ value cause the guided image to drift too far from the original generated distribution (i.e. too many guided denoising steps).

\subsubsection{Jeffrey guidance has a visible effect on Inception embedding distribution}
\label{sec:PCA_exp}

To go beyond quantitative FID effects of Jeffrey guidance, we now visualize the distributions of Inception embeddings for generated images and the target training data for the FFHQ dataset. This section therefore examines the distributions through a 2D PCA projection of the Inception embeddings for the combination $\delta=10, \lambda=30$ (which yield the best FID in \Cref{fig:jeffrey_fid}).

\begin{figure}[t]
    \centering

    \begin{subfigure}{0.32\linewidth}
        \centering
        \includegraphics[width=\linewidth]{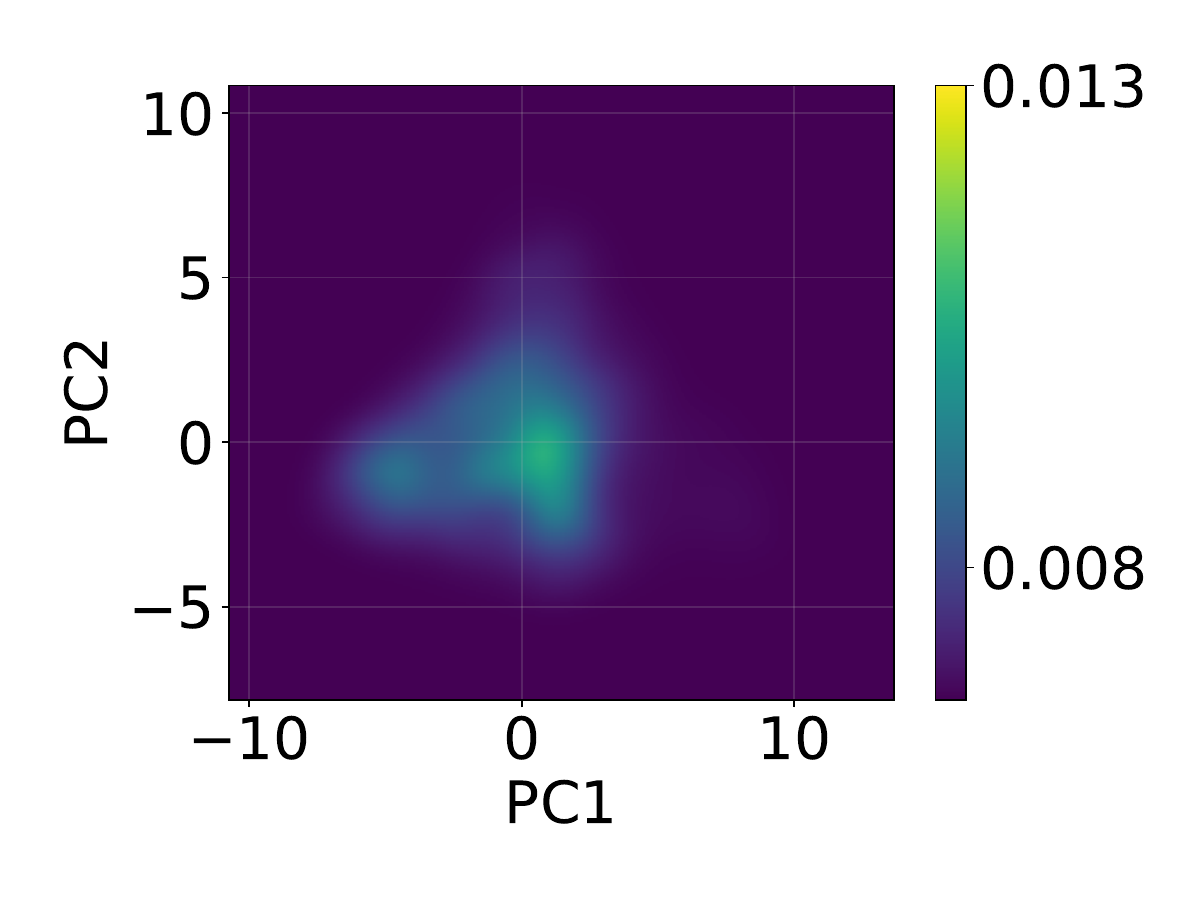}
        \caption{No guidance}
    \end{subfigure}
    \hfill
    \begin{subfigure}{0.32\linewidth}
        \centering
        \includegraphics[width=\linewidth]{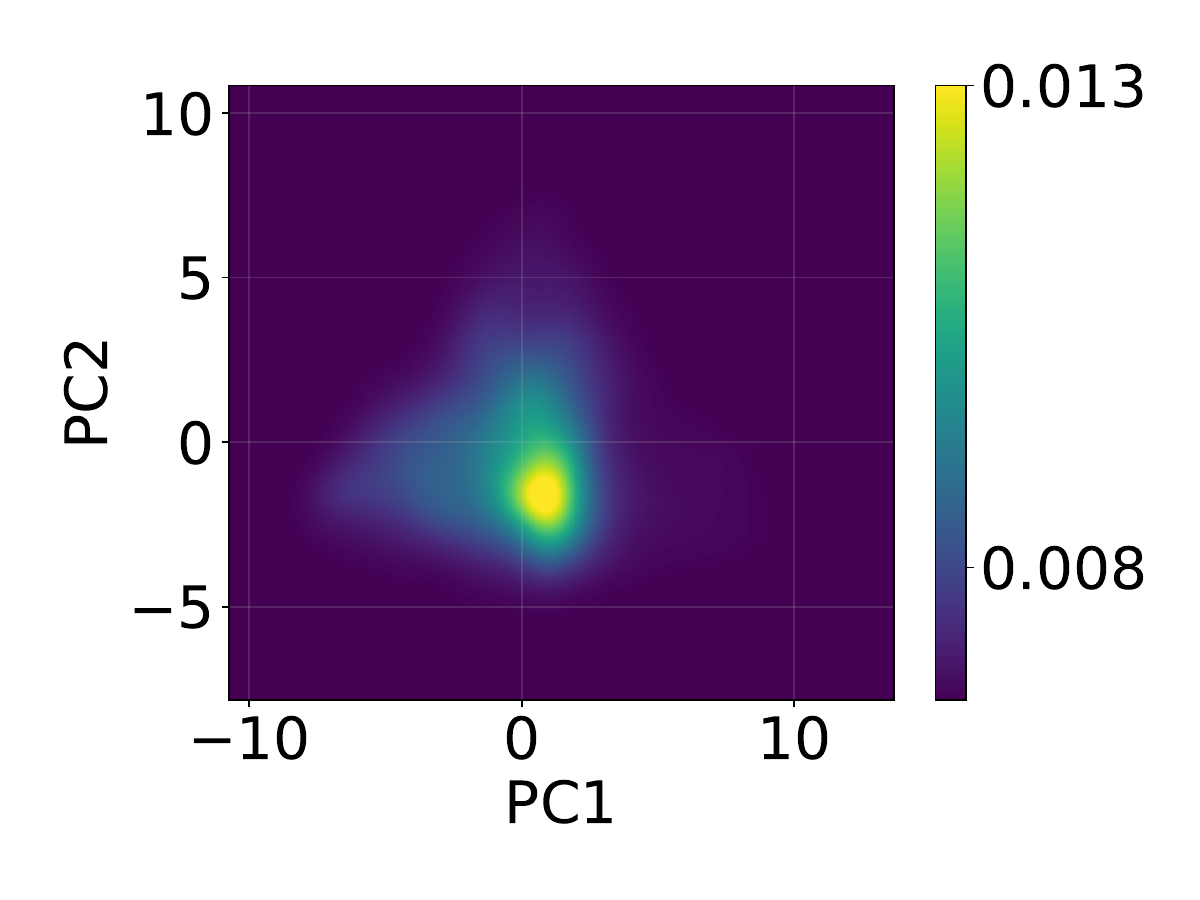}
        \caption{Jeffrey guidance}
    \end{subfigure}
    \hfill
    \begin{subfigure}{0.32\linewidth}
        \centering
        \includegraphics[width=\linewidth]{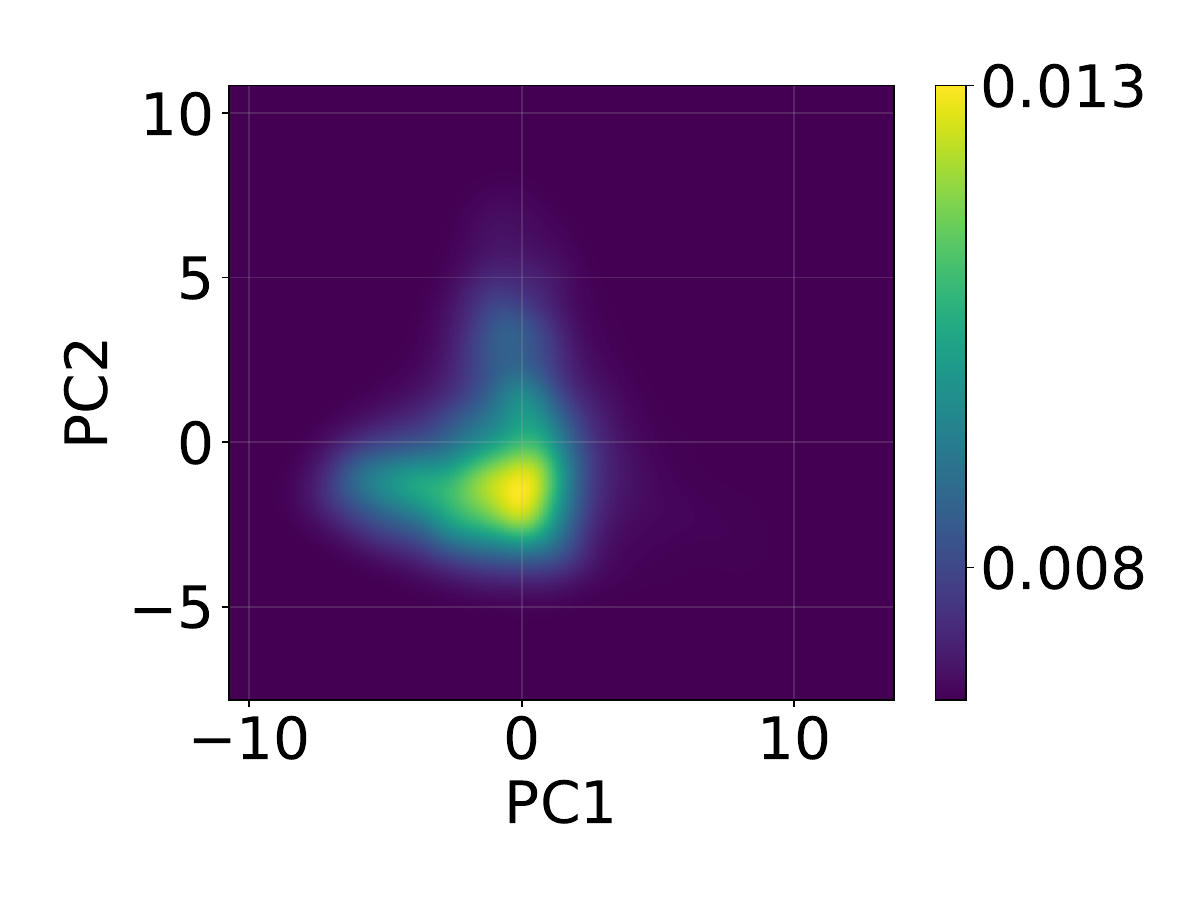}
        \caption{Test distribution}
    \end{subfigure}

    \caption{\textbf{PCA projection of Inception embeddings on FFHQ.} 
    Guided samples ($\lambda = 30.0$, $\delta = 10$) exhibit a closer alignment with the test distribution, as visible from the overall shape of the embedding distribution. See Appendix \Cref{fig:pca_appendix} for the contour map and \Cref{sec:experimental_details_appendix} for experimental details.}
    \label{fig:pca_heat}
\end{figure}

\Cref{fig:pca_heat} shows guided samples align more closely with the test distribution (not seen during training by the ratio estimator), confirming there is indeed a better match in feature space. Indeed, the distribution of Inception embeddings for images generated by an unguided model presents clear differences to that of the training data: multiple modes, different shape (see also \Cref{fig:pca_appendix}). By contrast, the guided distribution visually matches the characteristics of the targeted training distribution.

\subsection{Attribute distribution matching and decorrelation with Jeffrey guidance for fairness}
\label{sec:attribute_exp}

As discussed in \Cref{sec:fairness} and as we show in \Cref{sec:Parity_exp,sec:decorrelation_exp}, Jeffrey guidance can address important fairness problems like balancing attribute distributions and decorrelation of distributions. In all our experiments, we only use the CelebA-HQ dataset resized to 256 $\times$ 256 resolution, and follow \Cref{eq:guidance_scale_attribute} which implies classifiers to estimate the discrete distributions $p^\star(\mathbf{y})$ and $p(\mathbf{y})$ (see Appendix \Cref{subsec:distribution_matching_appendix,subsec:appendix_decorrelation_attributes}). We set $\delta = 1000$, which means guidance in all steps.


\subsubsection{Distribution matching for gender parity}\label{sec:Parity_exp}

\begin{figure}[t]
\centering

\begin{minipage}[t]{0.58\textwidth}
\vspace{0pt}
\centering

\captionsetup{type=table,position=top}
\captionof{table}{\textbf{Effect of guidance scale $\lambda$ on gender parity.} The best values are highlighted in bold for each metric and type of guidance. Here, $\mathbf{y}_{\mathrm{m}}$ denote the \emph{Male} attribute. The lowest FD is achieved for our Jeffrey guidance at $\lambda = 2.0$. For standard guidance, it is achieved at $\lambda=3.0$. We observe that, for our method, this improvement does not come at the expense of FID.}
\label{tab:gender_parity}

\scalebox{0.8}{
\small
\begin{tabular}{c|ccc|ccc}
\toprule
& \multicolumn{3}{c|}{Jeffrey guidance}
& \multicolumn{3}{c}{Standard guidance} \\
\cmidrule(lr){2-4} \cmidrule(lr){5-7}
$\lambda$
& $P(\mathbf{y}_{\mathrm{m}}{=}1)$
& FD $(\downarrow)$
& FID $(\downarrow)$
& $P(\mathbf{y}_{\mathrm{m}}{=}1)$
& FD $(\downarrow)$
& FID $(\downarrow)$ \\
\midrule

\rowcolor{blue!10}
$0.0$ & $0.296$ & $0.288$ & $22.30$
& $0.296$ & $0.288$ & $22.30$ \\

$0.5$ & $0.348$ & $0.214$ & $21.71$
& $0.319$ & $0.255$ & $22.94$ \\

$1.0$ & $0.411$ & $0.125$ & $\textbf{21.24}$
& $0.398$ & $0.142$ & $\textbf{22.34}$ \\

\cellcolor{orange!20}
$2.0$
& \cellcolor{orange!20}$\textbf{0.522}$
& \cellcolor{orange!20}$\textbf{0.032}$
& \cellcolor{orange!20}$21.61$
& $0.474$
& $0.037$
& $23.18$ \\

$3.0$ & $0.615$ & $0.162$ & $23.16$
& $\textbf{0.510}$ & $\textbf{0.014}$ & $24.33$ \\

$5.0$ & $0.767$ & $0.378$ & $28.28$
& $0.640$ & $0.198$ & $35.92$ \\

$8.0$ & $0.889$ & $0.551$ & $35.08$
& $0.781$ & $0.219$ & $56.31$ \\

\bottomrule
\end{tabular}
}

\vspace{-2.8cm}

\end{minipage}
\hfill
\begin{minipage}[t]{0.4\textwidth}
\vspace{0pt}
\centering

\includegraphics[width=\linewidth]{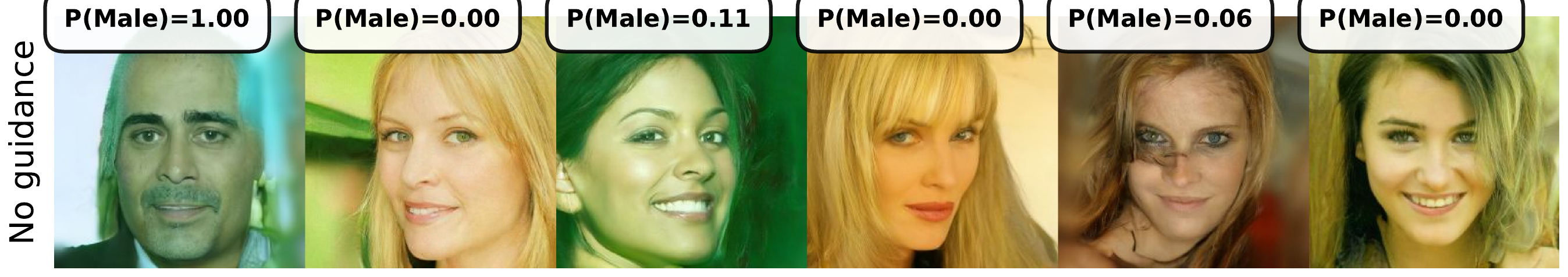}

\vspace{-0.1cm}

\includegraphics[width=\linewidth]{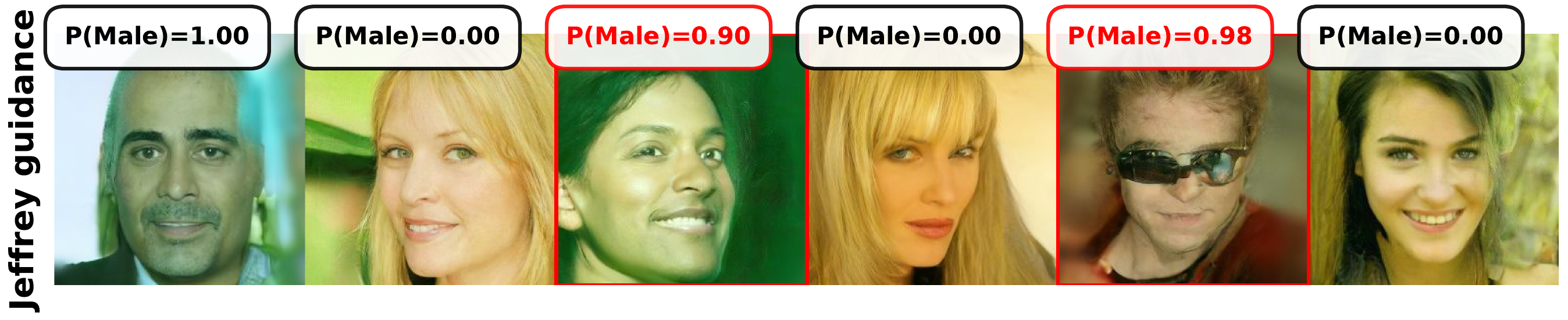}

\vspace{-0.1cm}

\includegraphics[width=\linewidth]{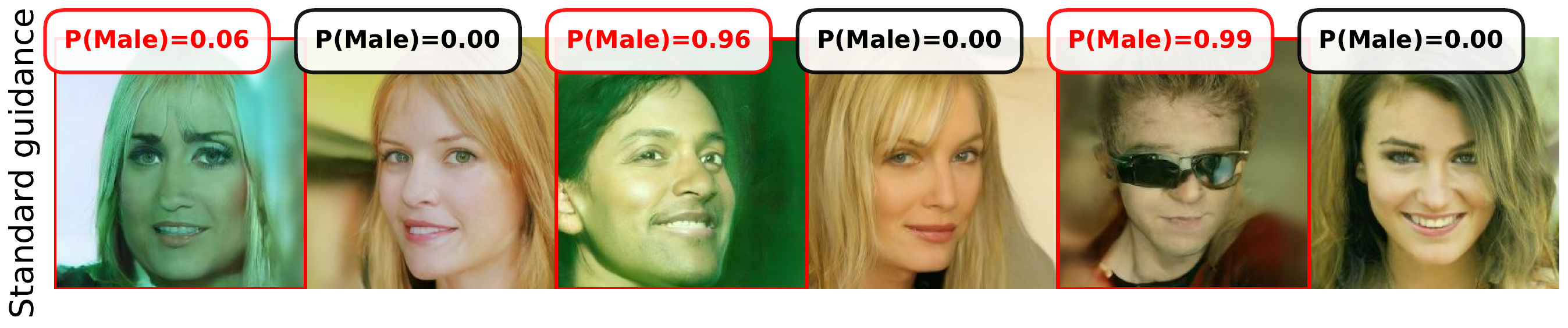}
\vspace{-0.15cm}

\captionsetup{type=figure,position=bottom}
\captionof{figure}{\textbf{Visualization of guidance for gender parity.} Jeffrey guidance selectively modifies samples (highlighted in red) leaving the  most confident samples essentially unchanged, while standard guidance modifies a lot more samples. More samples are available in \Cref{fig:balance_grids_extended}.}
\label{fig:balance_grids}

\end{minipage}

\end{figure}

We show as a proof of concept that Jeffrey guidance allows to steer the generated samples toward a balanced distribution of the \textit{Male} attribute (which is binary), represented through the feature $\mathbf{y}_{\mathrm{m}}$ with a careful choice of guidance strength. 
Following previous works on fairness \citep{choi2020fair,parihar2024balancing,2025debiasingdiff}, we evaluate distribution matching using the \textit{Fairness Discrepancy} (FD). This metric measures the deviation between a target distribution $\bar{p}$ and the average classifier predictions over generated samples: $
\mathrm{FD}
=\left\|
\bar{p}-\mathbb{E}_{\mathbf{x}\sim p_\theta(\mathbf{x})}
\big[c_\psi(\mathbf{x})\big]
\right\|_2$
where $\bar{p}$ denotes the target attribute distribution and $c_\psi(\mathbf{x})$ the softmax outputs of a pre-trained attribute classifier.

As measured by the FD in \Cref{tab:gender_parity}, moderate guidance strengths allow us to closely match the target distribution. In particular, using $\lambda=2.0$, the proportion of males shifts from $\approx 0.296$ to $\approx 0.522$, reducing the FD from $\approx 0.288$ to $\approx 0.032$. Larger guidance strengths eventually overcompensate and bias the distribution in the opposite direction, reaching a male proportion of $\approx 0.889$. Interestingly, achieving near-perfect balance does not translate into a significant degradation of FID, although excessively strong guidance increases it to $\approx 35.08$. This highlights a similar trade-off to standard class-conditional guidance \citep{dhariwal2021diffusion,ho2022classifier}, where the guidance strength must be chosen to balance sample quality and faithfulness to the desired target distribution. 

\Cref{tab:gender_parity} also shows the benefit of Jeffrey guidance over the natural baseline combining ancestral sampling with standard guidance (i.e., drawing the attribute class randomly and then guiding towards the drawn class). While simple class-conditional guidance does achieve a balanced distribution with FD $\approx 0.014$, it comes at the cost of higher FID valued to $\approx 24.33$. \Cref{fig:balance_grids} suggests the baseline approach leads to changing more samples' attribute compared to Jeffrey guidance. We believe this is because Jeffrey guidance continuously balances both class directions along the sampling trajectory (see \Cref{eq:guidance_scale_attribute}), whereas standard guidance pushes each sample entirely toward a single class.

\subsubsection{Decorrelation of Young and Male}\label{sec:decorrelation_exp}

\begin{table}[!t]
\centering
\caption{\textbf{Effect of guidance scale $\lambda$ on correlation $\varphi$.} $\mathbf{y}_{\mathrm{m}}$ and $\mathbf{y}_{\mathrm{y}}$ denote the \emph{Male} and \emph{Young} attributes, respectively. The very low correlation achieved (at $\lambda=3$, in bold) comes with relatively minor changes in the marginal distributions and a limited degradation of about one FID point. Notably, the most pronounced shift in the joint distribution is a reduction in the proportion of samples that are both female and young, which constitutes the dominant group in the unguided setting.}
\label{tab:guidance_correlation}
\scalebox{0.9}{
\small
\begin{tabular}{l|cccccccc}
\toprule
& & \multicolumn{2}{c}{Marginals} & \multicolumn{4}{c}{$P(\mathbf{y}_{\mathrm{m}},\mathbf{y}_{\mathrm{y}})$} & \\
\cmidrule(lr){3-4} \cmidrule(lr){5-8}
Guidance $\lambda$
& $\varphi(\mathbf{y}_{\mathrm{m}},\mathbf{y}_{\mathrm{y}})$
& $P(\mathbf{y}_{\mathrm{m}}{=}1)$
& $P(\mathbf{y}_{\mathrm{y}}{=}1)$
& $P(0,0)$
& $P(0,1)$
& $P(1,0)$
& $P(1,1)$
& FID $(\downarrow)$ \\
\midrule
\rowcolor{blue!10}
$0.0$ & $-0.541$ & $0.264$ & $0.822$ & $0.040$ & $0.696$ & $0.138$ & $0.126$ & $22.30$ \\
$1.0$ & $-0.351$ & $0.265$ & $0.810$ & $0.079$ & $0.656$ & $0.111$ & $0.154$ & $22.55$ \\
\rowcolor{orange!20}
$3.0$ & $\mathbf{0.005}$ & $0.274$ & $0.718$ & $0.206$ & $0.520$ & $0.076$ & $0.198$ & $23.59$ \\
$5.0$ & $0.283$ & $0.281$ & $0.587$ & $0.359$ & $0.360$ & $0.054$ & $0.228$ & $25.23$ \\
\bottomrule
\end{tabular}
}
\end{table}

As another illustration, we can steer generation toward decorrelated \emph{Male} $\mathbf{y}_{\mathrm{m}}$ and \emph{Young} $\mathbf{y}_{\mathrm{y}}$ attributes, measured by the Pearson correlation $\varphi$ coefficient between the attributes. 

\begin{wrapfigure}[16]{r}{0.5\textwidth}
\vspace{-.4cm}
    \centering
     \begin{subfigure}{\linewidth}
        \centering
        \includegraphics[width=0.95\linewidth]{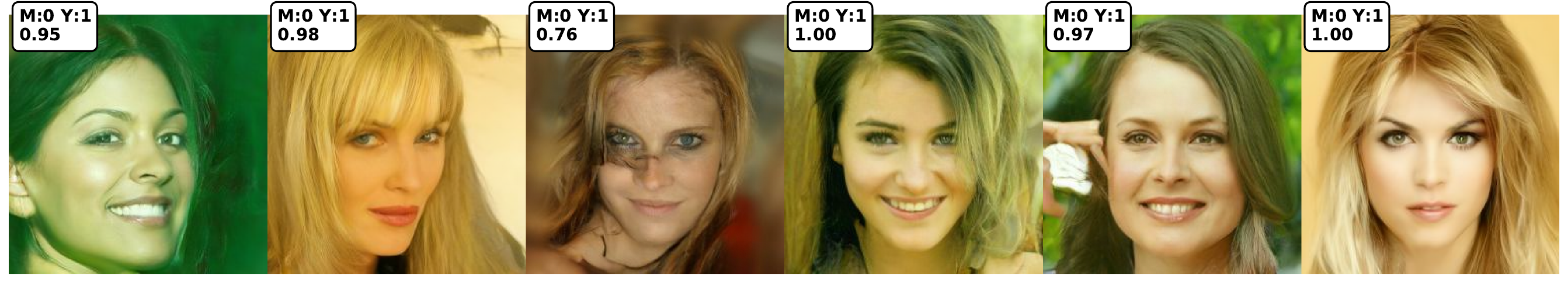}
    \end{subfigure}
    \vspace{-6mm}

    \begin{subfigure}{\linewidth}
        \centering
        \includegraphics[width=0.95\linewidth]{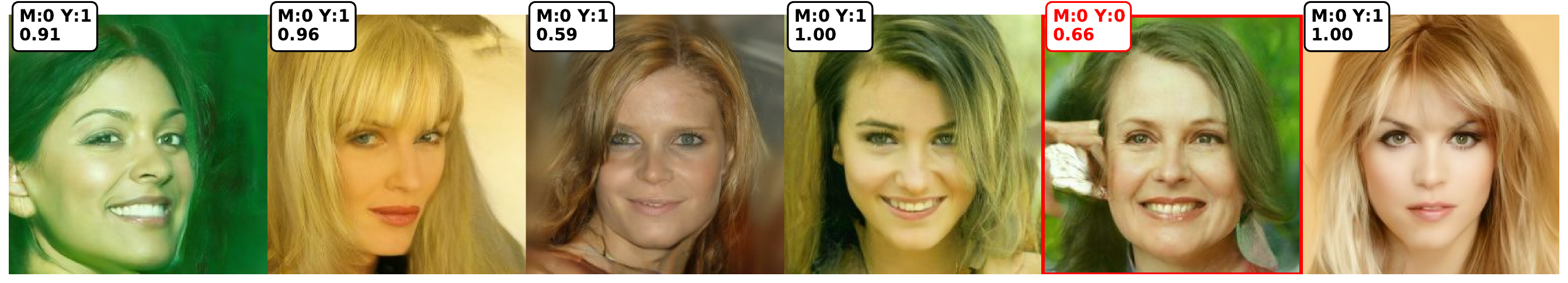}
    \end{subfigure}
    \vspace{-6mm}

    \begin{subfigure}{\linewidth}
        \centering
        \includegraphics[width=0.95\linewidth]{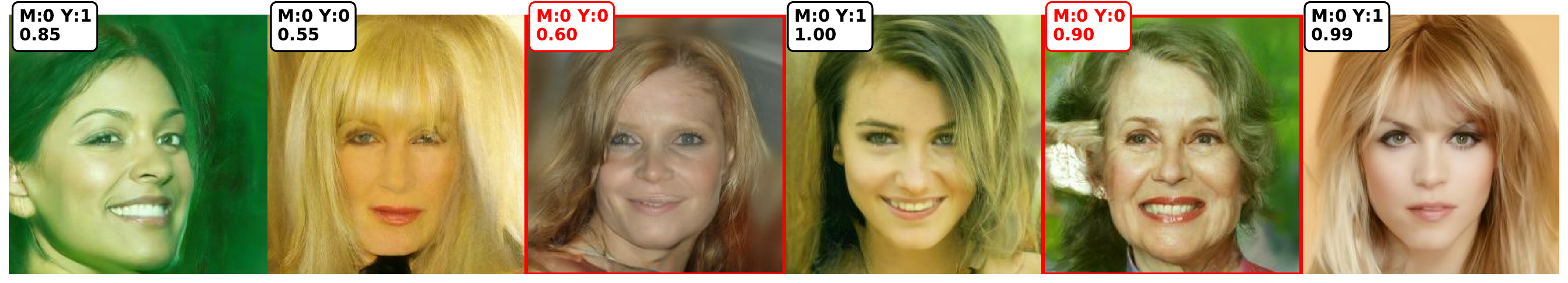}
    \end{subfigure}

    \caption{\textbf{Decorrelation of \textit{Male} and \textit{Young}: effect of $\lambda \in \{0.0,0.3,0.5\}$ (from top to bottom) on samples.} We highlight in \textcolor{red}{red} the transformed samples. The guidance modifies the joint distribution and, in particular, increases the probability of $\mathbf{y}_{\mathrm{m}} = 0$ and $\mathbf{y}_{\mathrm{y}} = 0$. More details in Fig.E\ref{fig:balance_grids_extended}.}
    \label{fig: decorrelation samples}
\end{wrapfigure}

\Cref{tab:guidance_correlation} shows Jeffrey guidance achieves near-zero correlation with $\lambda=3.0$ while inducing only minor changes in the marginals. Notably, this decorrelation is primarily driven by a reduction in the dominant joint mode corresponding to young female samples (see \Cref{fig: decorrelation samples}). Lower guidance strengths reduce the correlation but are insufficient to reach independence, whereas stronger guidance over-corrects, increasing correlation and distorting the marginals. We attribute the need for $\lambda > 1.0$ to approximations in our guidance procedure, for which a larger guidance scale might compensate. Importantly, this improvement comes at a limited cost of about one FID point at $\lambda = 3.0$. All these observations demonstrate that Jeffrey guidance can steer statistical relationships without explicitly prescribing the nature of individual variables, suggesting a principled framework for structural distribution control.

\section{Conclusion}

We introduce Jeffrey guidance, a novel formulation of diffusion guidance that takes advantage of Jeffrey's rule of conditioning to both extend classic guidance and introduce new previously inaccessible applications. As a first illustration, we show how Jeffrey guidance can match the distribution of generated images to a target distribution in some embedding space (e.g. match training data in Inception embedding space). Results show that this application of Jeffrey guidance both drastically lowers measured FID and also visually changes the distribution of generated images to match that of training images. As an additional application, we also demonstrate Jeffrey guidance can be used to make generated image attributes independent both theoretically and on a fairness issue of the CelebA-HQ dataset (decorrelating Young and Male attributes).

In the future, we believe Jeffrey guidance will open new avenues for diffusion model control such as memorization mitigation, domain adaptation of generated images to new contexts, or more efficient drug design. In this first proof of concept work, we have relied on a classifier to help estimate the density ratio at the heart of Jeffrey guidance. It is our hope that further work will remove this limitation similarly to the way standard guidance has evolved towards classifier-free guidance.

\paragraph{Broader impact}
While we hope this new expressive power will be used to beneficial ends such as memorization mitigation or improving fairness, we must acknowledge it will also further empower malicious actors to generate fake or nefarious outputs. Furthermore, even inference of diffusion models causes a significant pollution and energy burden at scale which our method will contribute to.

\begin{ack}

This work was supported by the French government through the France 2030 programme managed by the ANR. PAM was also supported by the French government, managed by the ANR, with the reference number “ANR-23-IACL-0001”, as well as RS with the reference number “ANR-15-IDEX-01”, and FP. JF acknowledge support from the Novo Nordisk Foundation through the Center for Basic Machine Learning Research in Life Science (MLLS, grant no.\@ NNF20OC0062606), the Independent Research Fund Denmark (grant nos.\@ 5334-00122B and 5334-00076B) and the Reinholdt W. Jorck og Hustrus Fond. We thank Thomas Hamelryck, Jesper Ferkinghoff-Borg, Zoubin Ghahramani, Dan Roy and Damien Garreau for valuable discussions over the years.
\end{ack}

\bibliographystyle{plainnat}
\bibliography{bibliography}


\appendix

\newpage

\section*{Appendix}

This appendix supplements our main contributions with additional insights, methodological clarifications,
and extended empirical results:
\begin{itemize}
\item Appendix~\ref{sec:appendix_deriving_guidance} provides a more detailed description of the attribute guidance formulation presented in \Cref{sec:attribute_exp}, including how the guidance terms and attribute distributions are estimated to compute the ratios.\item Appendix~\ref{sec:density_ratio_estimation} presents a more general approach for estimating density ratios using a binary classifier, corresponding to the setting considered in \Cref{sec:embedding_exp}.
\item Appendix~\ref{app: exactness} discusses a possible way to make the guidance formulation exact rather than approximate like our $\widehat{\mathbf{x}}_0$-based approach.
\item Appendix~\ref{sec:additional_embedding_figures} provides additional plots and figures to support \Cref{sec:embedding_exp}.
\item Appendix~\ref{sec:experimental_details_appendix} provides experimental details on training or inference.
\end{itemize}

\section{Attribute guidance: deriving the guidance terms with a classifier}\label{sec:appendix_deriving_guidance}

In this section, we explain how to obtain all the components of the guidance term. For the attribute guidance experiments, we will deal with discrete variables and formulate the guidance as
\begin{equation}\label{eq:guidance_scale_attribute}
\nabla_{\mathbf{x}_t} \log \tilde{p}^{\star}_t(\mathbf{x}_t)
= \nabla_{\mathbf{x}_t} \log p_t(\mathbf{x}_t)
+ \lambda\nabla_{\mathbf{x}_t} \log \sum_{\mathbf{a}_1,\dots,\mathbf{a}_n} p_{\bm{\theta}}(\mathbf{a}_1,\dots,\mathbf{a}_n\mid \widehat{\mathbf{x}}_0(\mathbf{x}_t,t))\frac{p^\star(\mathbf{a}_1,\dots,\mathbf{a}_n)}{p_{\bm{\theta}}(\mathbf{a}_1,\dots,\mathbf{a}_n)}.
\end{equation}
where $\lambda>0$ is the guidance scale.

\subsection{Distribution matching}\label{subsec:distribution_matching_appendix}

To evaluate distribution matching, we consider the gender parity task on CelebA-HQ $256\times256$, which contains a binary \emph{Male} attribute $\mathbf{a} \in \{0,1\}$. We consider the following additional guidance term
\begin{equation}\label{eq: gender balance score}
\nabla_{\mathbf{x}_t}\log \left(\frac{p_{\bm{\theta}}(1 \mid \widehat{\mathbf{x}}_0(\mathbf{x}_t,t))}{2p_{\bm{\theta}}(1)} + \frac{p_{\bm{\theta}}(0 \mid \widehat{\mathbf{x}}_0(\mathbf{x}_t,t))}{2p_{\bm{\theta}}(0)}\right).
\end{equation}
This time-dependent score balances both class directions at each time $t>0$, with weights dependent on the model distribution on the gender.

\paragraph{The model distribution $p_{\text{model}}(\mathbf{a})$.} We estimate this distribution by using the proportions of each label given by a high-accuracy classifier. To achieve this, we train one on the Celeba-HQ, the training set of the diffusion model (trained on 25k and validated on the rest 5k). Then we use this classifier on 10k generated samples. The attribute proportions are then saved upstream to the sampling during which we collect these to compute the sum.

\paragraph{The conditional probabilities $p_{\bm{\theta}}(\mathbf{a}\mid\widehat{\mathbf{x}}_0(\mathbf{x}_t,t))$.} These quantities are computed during the diffusion sampling process using the same classifier. Since the classifier is trained on clean samples, it is expected to be better suited to low-noise timesteps. Nevertheless, we observe that the guidance remains effective (\Cref{sec:Parity_exp,sec:decorrelation_exp}) even when applied at all timesteps.

\subsection{Decorrelation of attributes}\label{subsec:appendix_decorrelation_attributes}

We consider the decorrelation of \textit{Male} and \textit{Young} attributes, which are both binary. We consider the following additional guidance term
\begin{equation}
\nabla_{\mathbf{x}_t} \log \left( \sum_{\mathbf{a}_1,\mathbf{a}_2} p_{\bm{\theta}}(\mathbf{a}_1,\mathbf{a}_2\mid \widehat{\mathbf{x}}_0(\mathbf{x}_t,t))\frac{p_{\bm{\theta}}(\mathbf{a}_1) p_{\bm{\theta}}(\mathbf{a}_2)}{p_{\bm{\theta}}(\mathbf{a}_1,\mathbf{a_2})} \right).
\end{equation}

This time-dependent score balances joint probabilities across attributes.

\paragraph{The model distribution $p_{\bm{\theta}}(\mathbf{a}_1,\mathbf{a}_2)$ and the marginals.} We estimate all these components by using the proportions of labels (or couple of labels) given by a high-accuracy joint classifier: we train a classifier to predict $2^n = 4$ classes. Apart from this, the procedure is next the same as for distribution matching.

\paragraph{The conditional probabilities $p_{\bm{\theta}}(\mathbf{a}_1,\mathbf{a}_2\mid \widehat{\mathbf{x}}_0(\mathbf{x}_t,t))$.} The same classifier is used during the sampling procedure. Since the classifier is trained on clean samples, it is expected to be better suited to low-noise timesteps. Nevertheless, we observe that the guidance remains effective (\Cref{sec:Parity_exp,sec:decorrelation_exp}) even when applied at all timesteps.

\section{A quick note on density ratio estimation}

\label{sec:density_ratio_estimation}

In most interesting cases, the density ratio $\rho(y) = p^{\star}(y)/p(y)$, which lies at the core of Jeffrey’s update (\Cref{eq: deterministic jeffrey,eq: alternative jeffrey}) is unknown in practice. However, we are often given samples from both distributions, which allows us to estimate $\rho$. In simple settings, e.g. when $y$ is discrete, one can just estimate $p(y)$ and $p^\star(y)$ first and then compute the ratio. In more complex settings, in particular when $y$ is continuous and/or high-dimensional, estimating the ratio is more challenging. A standard approach, is to estimate it directly via probabilistic classification \citep{hastie2009elements,sugiyama2012density}. Specifically, one trains a binary classifier $\mathcal{D}_\phi$ to distinguish samples from $p^{\star}$ and $p$, assigning label $\ell=1$ to samples from $p^{\star}$ and $\ell=0$ to those from $p$. Using Bayes' rule, the density ratio can then be expressed in terms of the optimal classifier $\mathcal{D}^{\star}(y) = \mathbb{P}(\ell=1 \mid y)$ as\begin{equation}\label{eq: density ratio estimation}
\rho(y) = \frac{p^{\star}(y)}{p(y)} =  \frac{\mathbb{P}(y \mid \ell=1)}{\mathbb{P}(y \mid \ell=0)} = \frac{\mathbb{P}(\ell=1 \mid y)\,\mathbb{P}(\ell=0)}{\mathbb{P}(\ell=0 \mid y)\,\mathbb{P}(\ell=1)}  
   = \frac{\mathcal{D}^{\star}(y)}{1 - \mathcal{D}^{\star}(y)} \approx \frac{\mathcal{D}_\phi(y)}{1 - \mathcal{D}_\phi(y)},
\end{equation}if we assume balanced class priors $\mathbb{P}(\ell=0) = \mathbb{P}(\ell=1)$, which corresponds to equal numbers of samples from $p^{\star}$ and $p$. More complex estimation techniques exist (e.g. \citealp{choi2022density}) but we did not use them.

\section{On the exactness of the sampling procedure}\label{app: exactness}

In this section, we show that we rely in an approximation in our sampling procedure and that there exists an intermediate score formula that corresponds to the path that exactly the target distribution $p^\star(\mathbf{x})$.

For this discussion, we consider the deterministic Jeffrey formulation
\begin{equation}\label{eq: deterministic appendix}
p^\star(\mathbf{x})
=
p(\mathbf{x})\,\rho(u(\mathbf{x})),
\end{equation}
where $u:\mathcal{X}\rightarrow\mathbb{R}$ is a deterministic transformation. The score of $p^\star(\mathbf{x})$ is straightforward to derive:
\begin{equation}\label{eq: score energy appendix}
\nabla_{\mathbf{x}}\log p^\star(\mathbf{x})
=
\nabla_{\mathbf{x}}\log p(\mathbf{x})
-
\lambda\nabla_{\mathbf{x}}\log \rho(u(\mathbf{x})).
\end{equation} However, sampling from $p^\star(\mathbf{x})$ using reverse diffusion requires access to the \textbf{noisy }score $\nabla_{\mathbf{x}_t}\log p_t^\star(\mathbf{x}_t)$ for $t>0$, which is generally intractable. Indeed, the corresponding noisy distribution
\begin{equation}
p_t^\star(\mathbf{x}_t)
=
\int p_{t\mid 0}(\mathbf{x}_t\mid \mathbf{x}_0)\,p^\star(\mathbf{x}_0)\,\mathrm{d}\mathbf{x}_0
\end{equation}
is not directly available. 
Interestingly, \cite{lu2023contrastive} derives an exact expression for a general energy function $\mathcal{E}$ instead of log ratio, and shows that it involves the reverse transition kernel $p_{0\mid t}(\mathbf{x}_0\mid\mathbf{x}_t)$. Below we rewrite their theorem in our context. This result has also been demonstrated by \cite{denker2024deft} in the posterior sampling context \citep{chung2022diffusion}. 
\begin{proposition}[Exact sampling scheme through a time-dependent score, \citealp{lu2023contrastive}]
We define $p^{\star}_0(\mathbf{x}_0)$ the target distribution, where $p^{\star}_0 \coloneqq p^{\star}$. In the same way we rewrite $p_0 \coloneqq p$. For all $t>0$,
\begin{equation}\label{eq: exact guidance}
\nabla_{\mathbf{x}_t} \log p^{\star}_t(\mathbf{x}_t)
=
\nabla_{\mathbf{x}_t} \log p_t(\mathbf{x}_t)
+
\nabla_{\mathbf{x}_t} \log \int \rho(u(\mathbf{x}_0))\, p_{0\mid t}(\mathbf{x}_0 \mid \mathbf{x}_t)\, \mathrm{d}\mathbf{x}_0,
\end{equation}
and $p^{\star}_t$ admits the following expression, which does not require the explicit knowledge of $p^{\star}$:
\begin{equation}
p^{\star}_t(\mathbf{x}_t) \propto
\begin{cases}
\, p_0(\mathbf{x}_0)\, \rho(u(\mathbf{x}_0)) & \text{if } t = 0, \\
\, p_t(\mathbf{x}_t)\, \int \rho(u(\mathbf{x}_0))\, p_{0\mid t}(\mathbf{x}_0 \mid \mathbf{x}_t)\, \mathrm{d}\mathbf{x}_0 & \text{if } t > 0,
\end{cases}
\end{equation}
where $p_{0\mid t}(\mathbf{x}_0 \mid \mathbf{x}_t)$ denotes the backward transition kernel.

In particular, running the reverse-time diffusion process driven by the score $\nabla_{\mathbf{x}_t} \log p^{\star}_t(\mathbf{x}_t)$ yields samples $\mathbf{x}_0$ distributed according to $p^{\star}(\mathbf{x}_0)$.
\end{proposition}

\cite{lu2023contrastive} proposes to learn the additional guidance term through a contrastive energy prediction objective, which is shown to converge to the underlying energy function.

\paragraph{Difference with our approach.} In the spirit of several guidance methods, notably for inverse problems \citep{chung2022diffusion}, we instead directly apply guidance on the predicted sample $\widehat{\mathbf{x}}_0$:
\begin{equation}
\nabla_{\mathbf{x}_t}\log \rho\big(u(\widehat{\mathbf{x}}_0(\mathbf{x}_t,t))\big)
=
\nabla_{\mathbf{x}_t}\log \rho\left(u\left(\mathbb{E}_{p_{0|t}(\mathbf{x}_0|\mathbf{x}_t)}[\mathbf{x}_0]\right)\right).
\end{equation}
Using the posterior expression explicitly gives
\begin{equation}
\nabla_{\mathbf{x}_t}\log \rho\left(u\left(\int \mathbf{x}_0\, p_{0|t}(\mathbf{x}_0|\mathbf{x}_t)\, \mathrm{d}\mathbf{x}_0\right)\right).
\end{equation}
This introduces a bias due to the 
fact we exchanged the integral and the generally non-linear function $u$, especially when $u$ is complex (e.g. an Inception network). We refer the reader to \cite{lu2023contrastive} for a more detailed discussion of the induced gap, and a proof about the difference of approaches. Moreover, our approach generally requires to activate the guidance on a subset of the generation timesteps, based on the fact features do not emerge immediately in the generation but in a restricted interval of steps \citep{meng2021sdedit,li2024critical}. Restricting the timestep window has been shown to be a natural strategy when applying guidance on $\widehat{\mathbf{x}}_0$ \citep{chen2024towards,2025debiasingdiff}.

\subsection{Jeffrey guidance drastically improves FID without notable improvements to images}

\label{app:FID_analysis}

Now we show that while guidance can yield large improvements in FID through more optimal embeddings, its perceptual impact depends strongly on $\delta$: for small $\delta$, it induces almost no visible changes, whereas for larger $\delta$, it produces noticeable but often localized modifications. 

To investigate this, we compare guided and unguided samples generated from the same seed with deterministic sampling, and analyze both qualitative pairs and distance correlations. For small $\delta$ (\Cref{fig:best_pixel_samples_one_iteration}), even the largest embedding differences yield almost no perceptible changes, despite a large FID gap; similarly, pixel-based pairs exhibit negligible variations. In contrast, for larger $\delta$ (\Cref{fig:best_inception_samples_many_iterations,fig:best_pixel_samples_many_iterations}), differences become visible: guidance mainly alters background regions, with occasional improvements in perceptual quality (e.g., more realistic faces). These results are corroborated by the CMMD values reported in these figures, a well-known alternative perceptual metric \citep{jayasumana2024rethinking}, which respectively show slight degradation for small $\delta$ and improvement for large $\delta$.

Consistently, the correlation analysis (\Cref{fig:correlation_one_iteration,fig:correlation_many_iterations}) shows that embedding improvements alone can significantly reduce FID without visible changes, while guidance applied over multiple iterations increasingly translates these improvements into pixel-level modifications. This highlights that guidance first operates at a semantic level, and when applied across more denoising steps, can also lead to perceptual gains. 

Finally, samples with minimal embedding changes are typically already photo-realistic, with simple or smooth backgrounds (\Cref{fig:worst_samples_many_iterations}), as they reflect high-quality, typical patterns of the FFHQ distribution, leaving little room for further improvement through guidance.

\section{Embedding-optimal guidance for FID: additional figures}\label{sec:additional_embedding_figures}

\subsection{Error bars on FID}

\begin{figure}[h]
    \centering
    \includegraphics[width=1.0\linewidth]{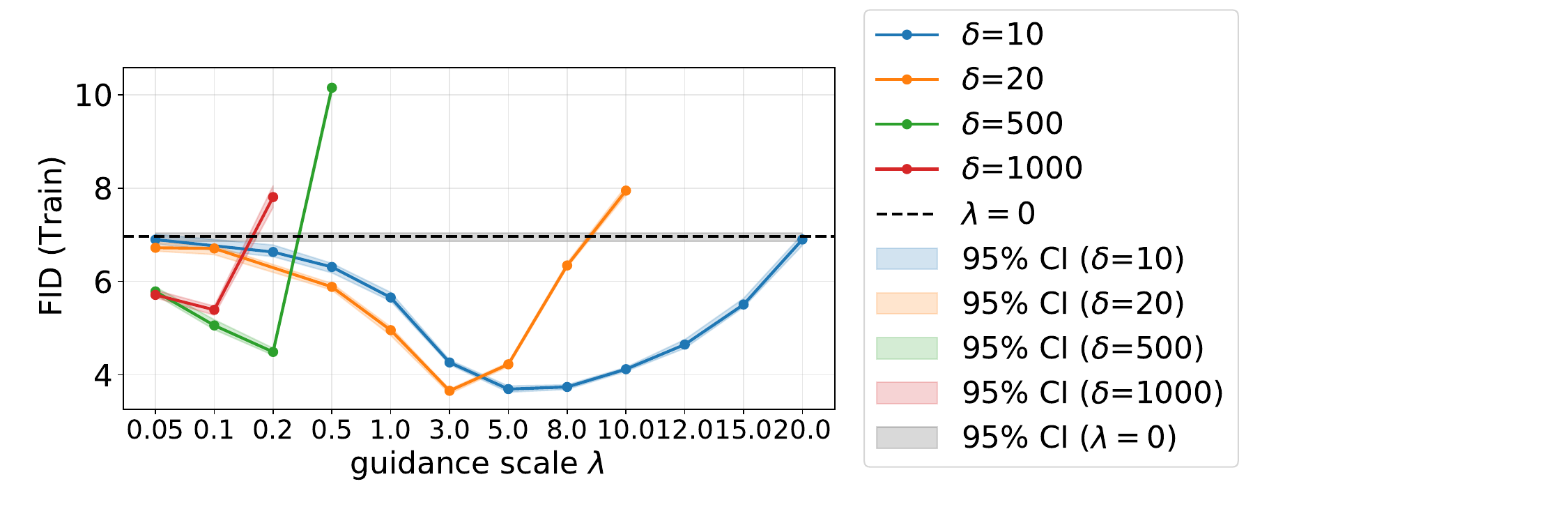}
    \caption{\textbf{FID as a function of the guidance scale $\lambda$ for different values of $\delta$, with $95\%$ confidence intervals.} In addition to the mean FID values reported in \Cref{fig:jeffrey_fid}, we display error bars to quantify the uncertainty of the metric estimation. We observe that the variability remains relatively small compared to the observed performance differences across guidance strengths.}
    \label{fig:jeffrey_fid_error_bars}
\end{figure}

To further assess in the reliability of the reported FID values in \Cref{fig:jeffrey_fid}, we estimate confidence intervals through repeated resampling (see \Cref{fig:jeffrey_fid_error_bars}). Since FID is computationally expensive to evaluate reliably, we approximate its variability by repeatedly computing the metric on randomly resampled subsets of generated and reference samples. More precisely, we sample $49$k generated samples and $49$k training samples with replacement, repeat the procedure $10$ times, and report empirical $95\%$ confidence intervals using the $2.5$ and $97.5$ percentiles. The resulting intervals, shown in \Cref{fig:jeffrey_fid_error_bars}, remain relatively small and support the significance of the observed trends.

\subsection{Standard guidance}

\begin{figure}[h]
    \centering
    \includegraphics[width=0.6\linewidth]{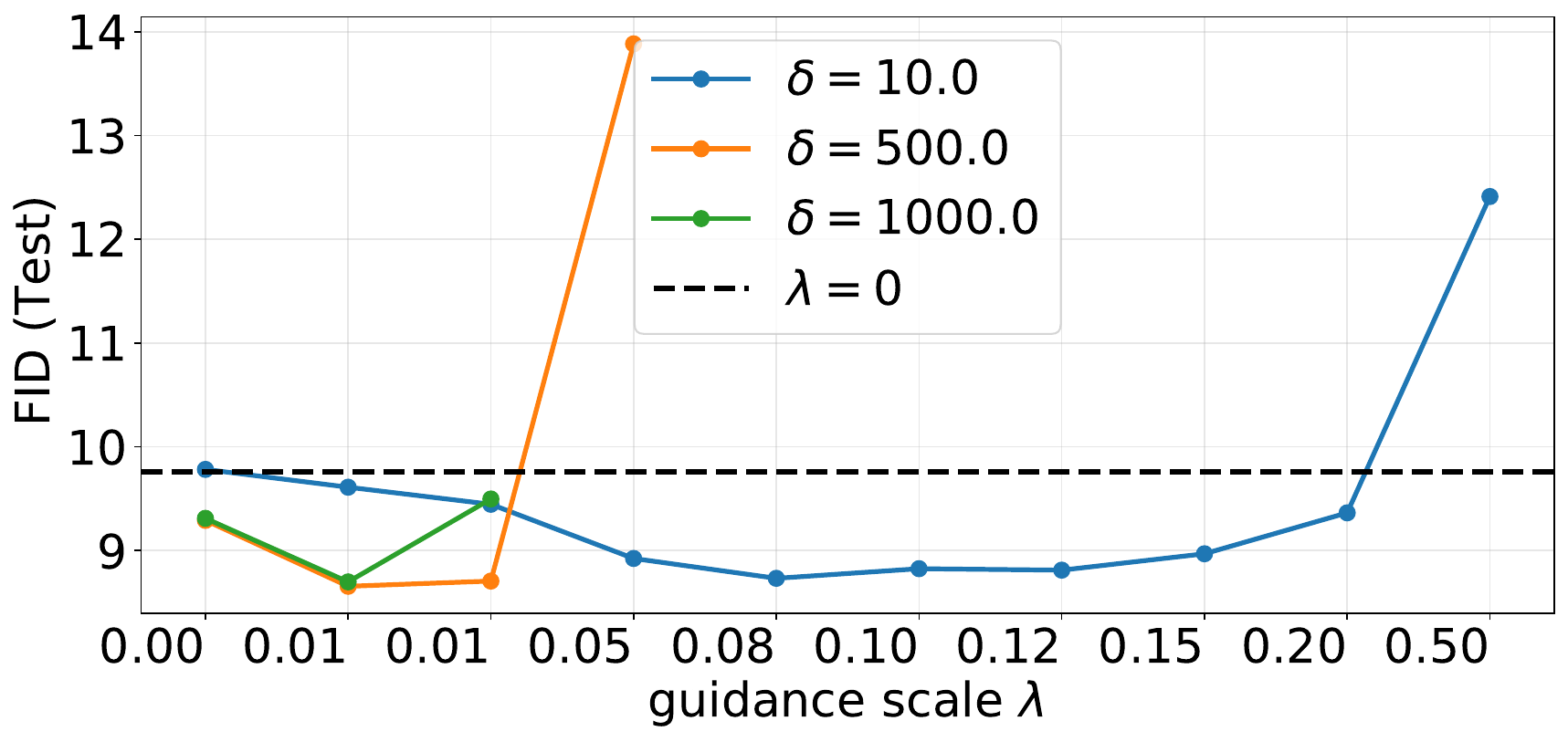}
    \caption{\textbf{CIFAR-10 FID as a function of the guidance scale $\lambda$ for different values of $\delta$ using ancestral sampling with guidance.} In this case, guidance also improves the FID on the test set, although the gap is not substantial.}
    \label{fig:jeffrey_baseline_fid}
\end{figure}

In \Cref{fig:jeffrey_baseline_fid}, standard guidance exhibits a trend similar to Jeffrey guidance on the CIFAR-10 test set. This proves that using the guidance term $\nabla_{\mathbf{x}_t}\log\mathcal{N}(\mathbf{y};u_{\text{in}}(\widehat{\mathbf{x}_0}(\mathbf{x}_t,t)),\boldsymbol{I})$ provides a relevant training-free baseline, improving the FID from $\approx 9.8$ by roughly one point. However, the improvement remains smaller than that achieved with Jeffrey guidance.  We note that $\lambda$ controls the variance of the Gaussian here.

\subsection{2D PCA embedding space: contour plot}

\begin{figure}[h]
    \centering
    \begin{subfigure}{0.4\linewidth}
        \centering
        \includegraphics[width=\linewidth]{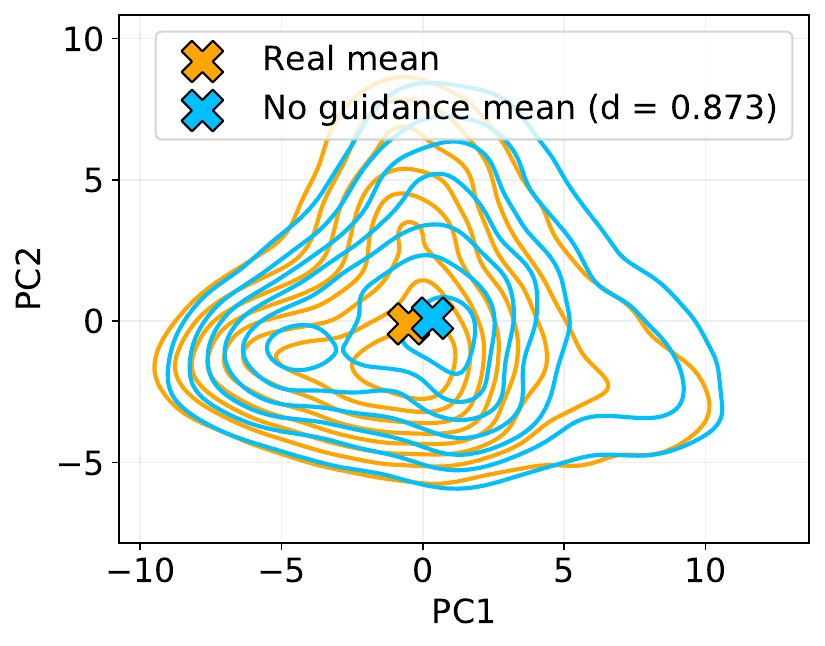}
        \caption{No guidance}
    \end{subfigure}%
    \begin{subfigure}{0.4\linewidth}
        \centering
        \includegraphics[width=\linewidth]{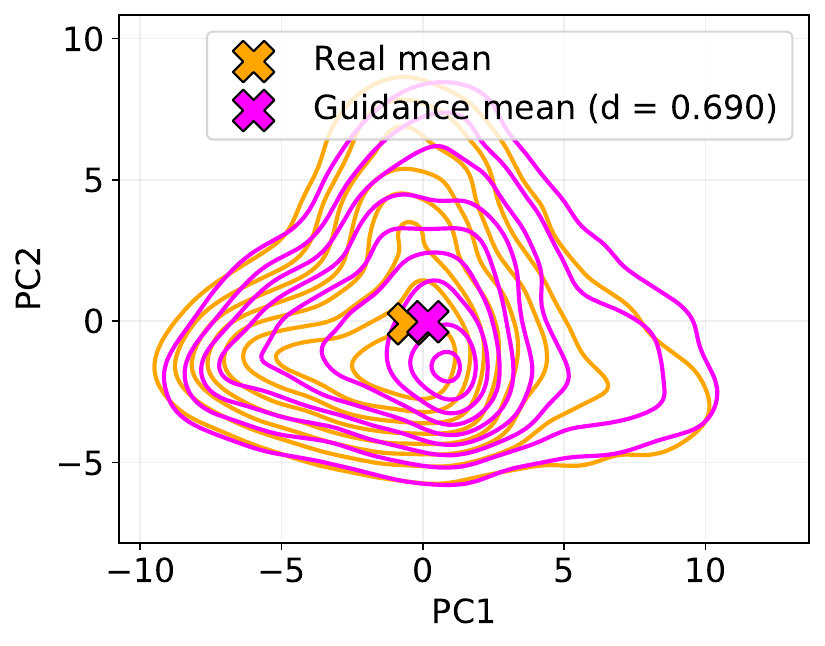}
        \caption{With guidance}
    \end{subfigure}

    \caption{
    \textbf{PCA projection of Inception embeddings on FFHQ.} 
    Guided samples ($\lambda = 30.0$, $\delta = 10$) exhibit a closer alignment with the test distribution, as visible from the overall shape of the embedding distribution and a reduced distance between empirical and reference means.
    }
    \label{fig:pca_appendix}
\end{figure}

We display in \Cref{fig:pca_appendix} an other version of \Cref{fig:pca_heat} through contour plots of the embedding distributions along with difference of means. 

\subsection{Relation ship between pixel changes and embeddings}

\begin{figure}[h]
    \centering
    \includegraphics[width=0.85\linewidth]{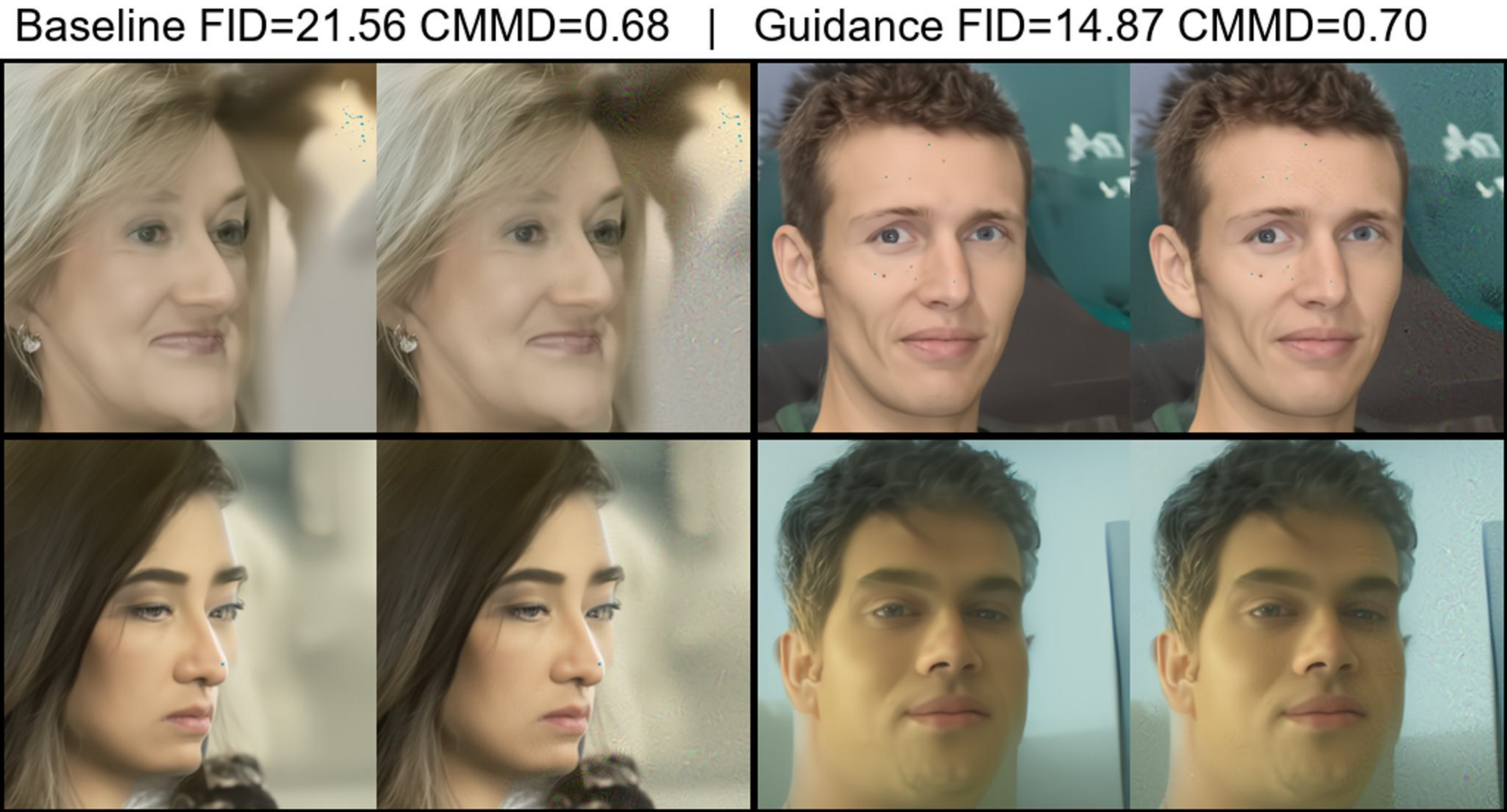}
    \caption{Top 4 pairs with the largest differences in pixel space between guided and unguided samples generated from the same seed ($\delta = 10$, $\lambda = 30.0$)}
    \label{fig:best_pixel_samples_one_iteration}
\end{figure}

\begin{figure}
    \centering
    \begin{subfigure}{0.4\linewidth}
        \centering
        \includegraphics[width=\linewidth]{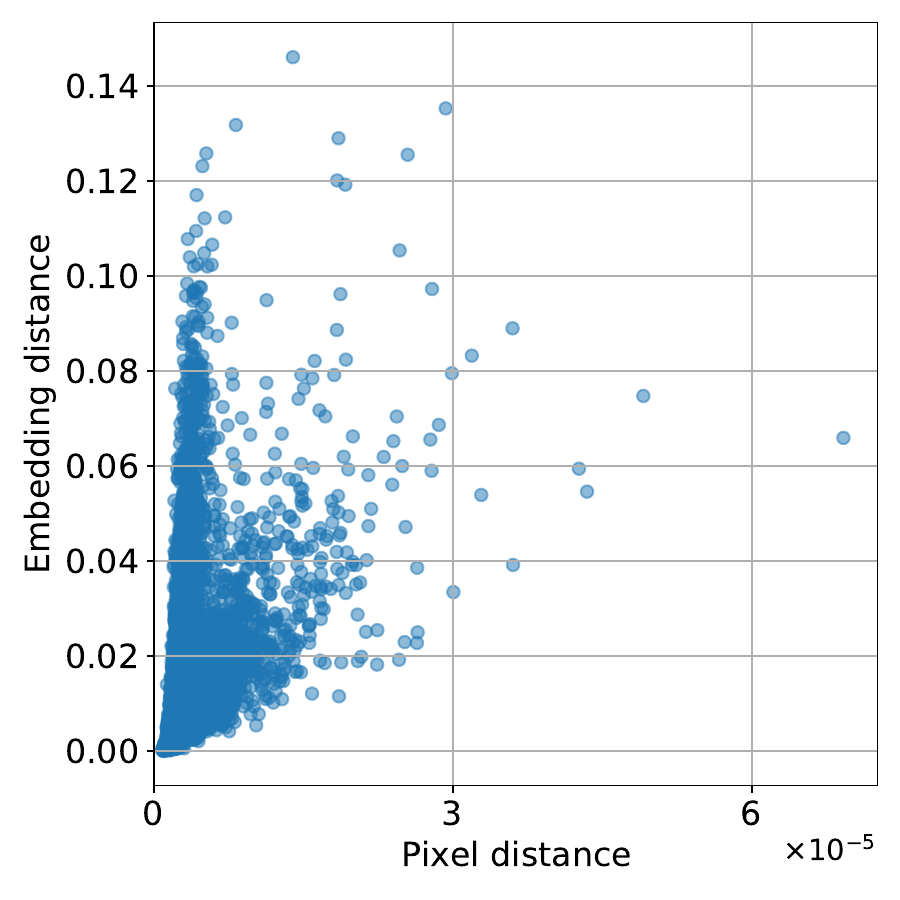}
        \caption{One iteration of guidance ($\delta = 10$)}
        \label{fig:correlation_one_iteration}
    \end{subfigure}
    \begin{subfigure}{0.4\linewidth}
        \centering
        \includegraphics[width=\linewidth]{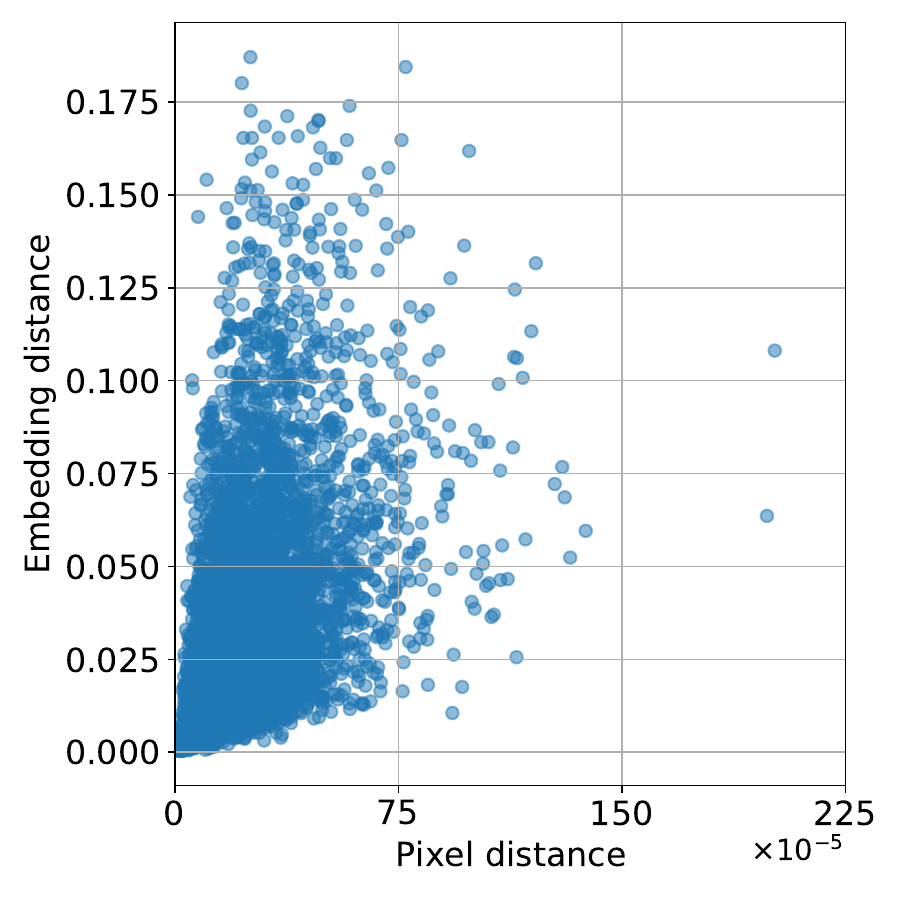}
        \caption{Multiple iterations of guidance ($\delta = 600$)}
        \label{fig:correlation_many_iterations}
    \end{subfigure}
    \caption{Correlation between embedding and pixel distances for different numbers of guidance iterations.}
    \label{fig:correlation_comparison}
\end{figure}

We display in \Cref{fig:best_inception_samples_many_iterations,fig:worst_samples_many_iterations,fig:best_pixel_samples_many_iterations,fig:best_pixel_samples_one_iteration,} the visual examples mentioned in \Cref{app:FID_analysis}. We also plot the correlation between embedding and pixel changes in \Cref{fig:correlation_comparison}.

\begin{figure}
    \centering
    \includegraphics[width=0.96\linewidth]{figs/comparison_baseline_guidance_embedding_many_iterations_big_jeffrey.pdf}
    \caption{Top 30 pairs with the largest differences in Inception embedding space between guided and unguided samples generated from the same seed ($\delta = 600$, $\lambda = 0.5$).}
    \label{fig:best_inception_samples_many_iterations}
\end{figure}
\begin{figure}
    \centering
    \includegraphics[width=0.96\linewidth]{figs/comparison_baseline_guidance_pixel_many_iterations_big_jeffrey.pdf}
    \caption{Top 30 pairs with the largest differences in pixel space between guided and unguided samples generated from the same seed ($\delta = 600$, $\lambda = 0.5$)}
    \label{fig:best_pixel_samples_many_iterations}
\end{figure}

\begin{figure}
    \centering
    \includegraphics[width=0.96\linewidth]{figs/comparison_baseline_guidance_embedding_many_iterations_big_bottom_jeffrey.pdf}
    \caption{Bottom 30 pairs with the largest differences in Inception embedding space between guided and unguided samples generated from the same seed ($\delta = 600$, $\lambda = 0.5$)}
    \label{fig:worst_samples_many_iterations}
\end{figure}

\section{Attribute guidance for fairness: additional figures} 

\begin{figure}[h]
    \centering
    \includegraphics[width=1\linewidth]{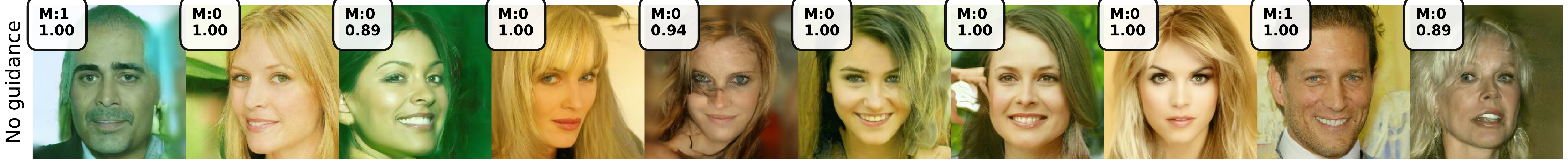}
    \vspace{-1.3em}

    \includegraphics[width=1\linewidth]{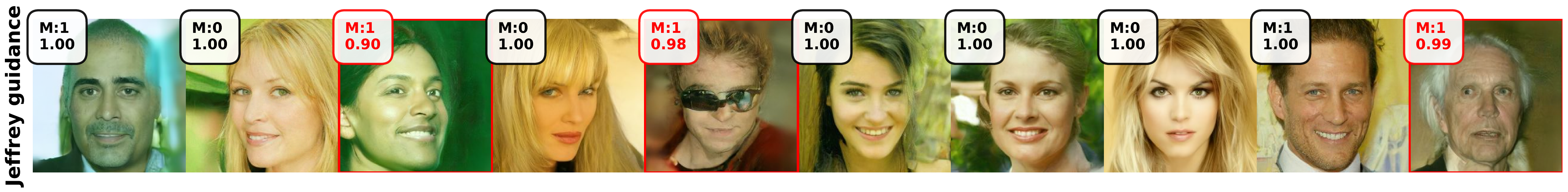}
    \vspace{-1.3em}

    \includegraphics[width=1\linewidth]{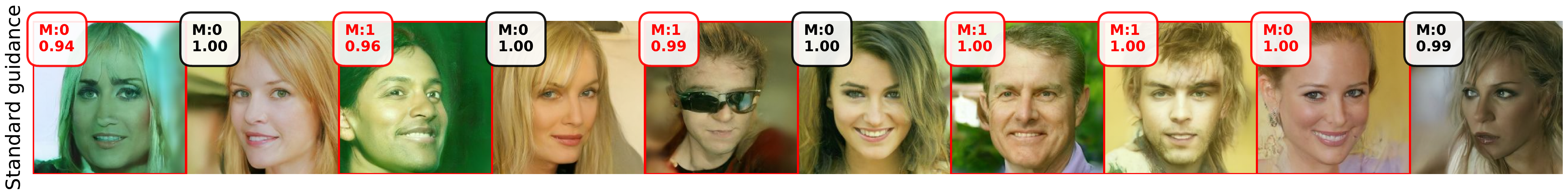}

    \caption{\textbf{Visualization of distribution matching for gender balance.}
    For each guidance strategy, we adopt the best $\lambda$ from \Cref{tab:guidance_correlation}, e.g., the one achieving the lowest FD. The value of M $\in \{0,1\}$ corresponds to the predicted label ($M=1$ for \textit{Male}) and below each value is reported the class probability.
    We observe that Jeffrey's guidance is selective regarding modifications (highlighted in red) and leaves the highest-confidence samples (e.g., with predicted probability 1.00) essentially unchanged, while standard guidance modifies a much larger fraction of samples.}
    \label{fig:balance_grids_extended}
\end{figure}

In \Cref{fig:balance_grids_extended}, we provide a large amount of samples to illustrate the distribution shift induced by decorrelation, with respect to the guidance strength $\lambda>0$.

\section{Experimental details}\label{sec:experimental_details_appendix}

In this section we report the experimental details.

\subsection{Matching embedding distributions with Jeffrey guidance}

\paragraph{Diffusion model.} For FFHQ-256, we use the ADM architecture from OpenAI’s official repository, which includes several improvements over the previous U-Net design \citep{dhariwal2021diffusion}. The model is trained using the DDPM objective \citep{ho2020denoisingdiffusionprobabilisticmodels}. We follow the repository’s recommendation for the learning rate (1e-4), while the batch size (4) and total number of training iterations (10M) are chosen based on our computational resources. No dropout is used in this setup. At sampling time, we use the DDIM sampler with $100$ steps with an entropy level $\eta=0.2$ (see \citet{song2020denoising} for a definition) by default.

For CIFAR-10, we adopted the checkpoint and architecture available in https://github.com/ermongroup/ddim, which implements DDIM \citep{ho2020denoisingdiffusionprobabilisticmodels}.

\paragraph{The density ratio estimation.} We used a simple linear classifier consisting of a single fully connected layer. The model was trained with cross-entropy loss for $15$ epochs using a learning rate of $10^{-3}$. 

\paragraph{KDE plot.} Kernel density estimates \Cref{fig:pca_heat,fig:pca_appendix} were computed in the two-dimensional PCA space of Inception embeddings using the Gaussian kernel implementation \texttt{gaussian\_kde} from \texttt{scipy.stats} with Scott's bandwidth rule. To improve visualization of low-density regions, we applied a power-law normalization with exponent $3.0$. 

\subsection{Attribute distribution matching and decorrelation with Jeffrey guidance for fairness}

\paragraph{Diffusion model.} For CelebA-HQ, we relied on the publicly available pre-trained diffusion model \url{google/ddpm-celebahq-256} from Hugging Face.

\paragraph{The density ratio estimation.} We relied on a ResNet initialized with standard ImageNet pre-trained weights from \texttt{torchvision} and fine-tuned it for attribute classification in order to estimate the proportions required for the density ratios. Early stopping was performed using a validation set.

\subsection{Compute costs for the experiments}
\label{sec:compute}

The main computational cost of our experiments comes from evaluation rather than training. In particular, reliable FID estimation requires generating tens of thousands of samples for each configuration, which largely dominates the overall runtime. Since we explore multiple guidance strengths $\lambda$, diffusion windows $\delta$, and guidance strategies, the total number of generated samples becomes substantial. Additional computation have been made on larger $\lambda$ but since the reported values of FID were too large we did not report them.

To make these evaluations tractable, experiments were executed in parallel on GPUs within a high-performance computing cluster over one to two days of computation (for example for an FID plot in \Cref{fig:jeffrey_fid}). We relied on publicly available pre-trained diffusion models for all considered datasets, while attribute classifiers were obtained by fine-tuning pre-trained backbones on the corresponding tasks.

\subsubsection{Python packages and repositories used}
\label{subsec:packages}

We used the following libraries for general scientific programming:
\begin{itemize}
    \item sklearn \citep{pedregosa2011scikit}, license: BSD 3
    \item Tensorflow \citep{tensorflow2015-whitepaper}, including Tensorflow-GAN, license: Apache 2.0
    \item numpy \citep{harris2020array}, license: BSD
    \item matplotlib \citep{Hunter:2007}, custom license, see \url{https://matplotlib.org/stable/project/license.html}
    \item pandas \citep{reback2020pandas}, license: BSD 3
    \item Pytorch \citep{ansel2024pytorch}, including Torchvision, license: BSD 3.
\end{itemize}
We also relied on the following repositories:
\begin{itemize}
    \item \url{https://github.com/Kaggle/kagglehub}, license: Apache 2.0
    \item \url{https://github.com/openai/guided-diffusion}, license: MIT
    \item \url{https://github.com/mseitzer/pytorch-fid}, license: Apache 2.0
    \item \url{https://github.com/openai/improved-diffusion}, license: MIT
    \item \url{https://github.com/ermongroup/ddim}, license: MIT.
    \item \url{https://huggingface.co/google/ddpm-celebahq-256}, license: Apache 2.0
\end{itemize}

\newpage

\section*{NeurIPS Paper Checklist}

\begin{enumerate}

\item {\bf Claims}
    \item[] Question: Do the main claims made in the abstract and introduction accurately reflect the paper's contributions and scope?
    \item[] Answer: \answerYes{} 
    \item[] Justification: The main claims of our paper are outlined up in the three bullet points at the end of out introduction. The first contribution (Jeffrey guidance itself) is provided and discussed in \Cref{sec:jg_density}. The second point is discussed theoretically in \Cref{sec:embedding} and experimentally in \Cref{sec:embedding_exp} (with the proper caveats around the FID gains). The third point is discussed theoretically in \Cref{sec:fairness} and experimentally in \Cref{sec:attribute_exp}.

\item {\bf Limitations}
    \item[] Question: Does the paper discuss the limitations of the work performed by the authors?
    \item[] Answer: \answerYes{} 
    \item[] Justification: We discuss the approximate nature of our guidance implementation (due to $\hat{x}_0$), contextualize the relative meaningfulness of our FID gains, name the exact datasets used in the study and acknowledge the cost of using high $\delta$ values. Furthermore, we highlight in the conclusion the limited scope of this first proof of concept like the reliance on a classifier model.

\item {\bf Theory assumptions and proofs}
    \item[] Question: For each theoretical result, does the paper provide the full set of assumptions and a complete (and correct) proof?
    \item[] Answer:  \answerYes{} 
    \item[] Justification: We use several results, and reference them in the text, together with their assumptions. There are no new theorems, just somewhat straightforward derivations.

    \item {\bf Experimental result reproducibility}
    \item[] Question: Does the paper fully disclose all the information needed to reproduce the main experimental results of the paper to the extent that it affects the main claims and/or conclusions of the paper (regardless of whether the code and data are provided or not)?
    \item[] Answer: \answerYes{} 
    \item[] Justification: We have indicated, to the best of our ability, all experimental design choices, hyperparameters used for computations, procedures used for each experiment, datasets and models used in the main text of this paper. When this was not possible, we provided additional detais in the supplementary material. The code is not made available now but would be if the paper is accepted.

\item {\bf Open access to data and code}
    \item[] Question: Does the paper provide open access to the data and code, with sufficient instructions to faithfully reproduce the main experimental results, as described in supplemental material?
    \item[] Answer: \answerNo{} 
    \item[] Justification: We will provide the code in open-source repository if the paper is accepted.

\item {\bf Experimental setting/details}
    \item[] Question: Does the paper specify all the training and test details (e.g., data splits, hyperparameters, how they were chosen, type of optimizer) necessary to understand the results?
    \item[] Answer: \answerYes{} 
    \item[] Justification: Experimental details are provided both in the main text (Section \ref{sec:experiments}) and in more details in the Appendix (\Cref{sec:experimental_details_appendix})

\item {\bf Experiment statistical significance}
    \item[] Question: Does the paper report error bars suitably and correctly defined or other appropriate information about the statistical significance of the experiments?
    \item[] Answer: \answerYes{} 
    \item[] Justification: Experiments are quite costly because we use somewhat large generative models. For this reason, we only provide error bars in some experiments: \Cref{fig:jeffrey_fid_error_bars} shows error bars on FID evaluated on multiple configurations.

\item {\bf Experiments compute resources}
    \item[] Question: For each experiment, does the paper provide sufficient information on the computer resources (type of compute workers, memory, time of execution) needed to reproduce the experiments?
    \item[] Answer: \answerYes{} 
    \item[] Justification: We discuss the computational resources used for this paper (in terms of orders of magnitude) in the supplemental material \Cref{sec:compute} 
    
\item {\bf Code of ethics}
    \item[] Question: Does the research conducted in the paper conform, in every respect, with the NeurIPS Code of Ethics \url{https://neurips.cc/public/EthicsGuidelines}?
    \item[] Answer: \answerYes{} 
    \item[] Justification: We have reviewed the code of ethics and verified that the paper does not breach any of the issues outlined in the code. In particular, we make sure to provide details for reproducibility, and discuss societal impacts in the main text.

\item {\bf Broader impacts}
    \item[] Question: Does the paper discuss both potential positive societal impacts and negative societal impacts of the work performed?
    \item[] Answer: \answerYes{} 
    \item[] Justification: We provide a broader impact statement at the end of our conclusion outlining the main positive and negative impacts we see with our work.
    
\item {\bf Safeguards}
    \item[] Question: Does the paper describe safeguards that have been put in place for responsible release of data or models that have a high risk for misuse (e.g., pre-trained language models, image generators, or scraped datasets)?
    \item[] Answer: \answerNA{} 
    \item[] Justification: The models trained for this paper are just standard diffusion models trained on classic image generation datasets: similar models are widespread and available. No dataset has been developed for this paper.

\item {\bf Licenses for existing assets}
    \item[] Question: Are the creators or original owners of assets (e.g., code, data, models), used in the paper, properly credited and are the license and terms of use explicitly mentioned and properly respected?
    \item[] Answer:  \answerYes{} 
    \item[] Justification: We provide a list of the assets we used (together with licenses and citations) in Appendix \ref{subsec:packages}.

\item {\bf New assets}
    \item[] Question: Are new assets introduced in the paper well documented and is the documentation provided alongside the assets?
    \item[] Answer:  \answerNA{} 
    \item[] Justification: The paper does not release new assets. A repository with the code will be published if the paper is accepted.

\item {\bf Crowdsourcing and research with human subjects}
    \item[] Question: For crowdsourcing experiments and research with human subjects, does the paper include the full text of instructions given to participants and screenshots, if applicable, as well as details about compensation (if any)? 
    \item[] Answer: \answerNA{} 
    \item[] Justification: The paper does not involve crowdsourcing nor research with human subjects.

\item {\bf Institutional review board (IRB) approvals or equivalent for research with human subjects}
    \item[] Question: Does the paper describe potential risks incurred by study participants, whether such risks were disclosed to the subjects, and whether Institutional Review Board (IRB) approvals (or an equivalent approval/review based on the requirements of your country or institution) were obtained?
    \item[] Answer: \answerNA{} 
    \item[] Justification: The paper does not involve crowdsourcing nor research with human subjects.

\item {\bf Declaration of LLM usage}
    \item[] Question: Does the paper describe the usage of LLMs if it is an important, original, or non-standard component of the core methods in this research? Note that if the LLM is used only for writing, editing, or formatting purposes and does \emph{not} impact the core methodology, scientific rigor, or originality of the research, declaration is not required.
    \item[] Answer: \answerNA{} 
    \item[] Justification: We did not use LLMs for any important aspect of this work.

\end{enumerate}

\end{document}